\documentclass[lettersize,journal]{IEEEtran}
\usepackage{amsmath,amssymb,amsfonts}
\usepackage{array}
\usepackage[caption=false,font=scriptsize,labelfont=rm,textfont=rm]{subfig}
\usepackage{textcomp}
\usepackage{stfloats}
\usepackage{url}
\usepackage{verbatim}
\usepackage{graphicx}
\usepackage{cite}
\usepackage{bm}
\usepackage{xcolor}
\usepackage{pifont}
\usepackage{fancyhdr}
\usepackage{multirow}
\usepackage[ruled,linesnumbered]{algorithm2e}
\usepackage{booktabs}
\usepackage{fontawesome}
\usepackage{epstopdf}
\newcommand{\cmark}{\ding{51}}%
\newcommand{\xmark}{\ding{55}}
\newcommand{\CBmark}{\faCheckSquareO}
\def\BibTeX{{\rm B\kern-.05em{\sc i\kern-.025em b}\kern-.08em
    T\kern-.1667em\lower.7ex\hbox{E}\kern-.125emX}}
\begin{document}
\title{A Multi-scale Generalized Shrinkage Threshold Network for Image Blind Deblurring in Remote Sensing}
\author{Yujie Feng, Yin Yang, Xiaohong Fan, Zhengpeng Zhang, and Jianping Zhang
\thanks{This work was supported by the National Natural Science Foundation of China Project (12071402, 12261131501), the Education Bureau of Hunan Province, China (22A0119), the National Key Research and Development Program of China (2020YFA0713503), the Natural Science Foundation of Hunan Province (2020JJ2027, 2020ZYT003, 2023GK2029), the Project of Scientific Research Fund of the Hunan Provincial Science and Technology Department (2022RC3022), the Postgraduate Scientific Research Innovation Project of Hunan Province (CX20220633) and Postgraduate Scientific Research Innovation Project of Xiangtan University (XDCX2022Y060), China.}
\thanks{Corresponding author: J. Zhang (e-mail: jpzhang@xtu.edu.cn).}
\thanks{Y. Feng and X. Fan are with the School of Mathematics and Computational Science, Xiangtan University, Hunan Key Laboratory for Computation and Simulation in Science and Engineering, Key Laboratory for Intelligent Computing and Information Processing of the Ministry of Education, Xiangtan 411105, China.}
\thanks{Y. Yang and J. Zhang are with the School of Mathematics and Computational Science, Xiangtan University, National Center for Applied Mathematics in Hunan, Hunan International Scientific and Technological Innovation Cooperation Base of Computational Science, Xiangtan 411105, China.}
\thanks{Z. Zhang is with the School of Automation and Electronic Information, Xiangtan University, Xiangtan 411105, China.}
}

\markboth{IEEE Transactions on Geoscience and Remote Sensing,~Early Access, No.~xx, Feb.~2024, DOI: 10.1109/TGRS.2024.3368760}%
{How to Use the IEEEtran \LaTeX \ Templates}

\maketitle

\begin{abstract}
Remote sensing images are essential for many applications of the earth's sciences, but their quality can usually be degraded due to limitations in sensor technology and complex imaging environments. To address this, various remote sensing image deblurring methods have been developed to restore sharp and high-quality images from degraded observational data. However, most traditional model-based deblurring methods usually require predefined {hand-crafted} prior assumptions, which are difficult to handle in complex applications. On the other hand, deep learning-based deblurring methods are often considered as black boxes, lacking transparency and interpretability. In this work, we propose a new blind deblurring learning framework that utilizes alternating iterations of shrinkage thresholds. This framework involves updating blurring kernels and images, with a theoretical foundation in network design. Additionally, we propose a learnable blur kernel proximal mapping module to improve the accuracy of the blur kernel reconstruction. Furthermore, we propose a deep proximal mapping module in the image domain, which combines a generalized shrinkage threshold with a multi-scale prior feature extraction block. This module also incorporates an attention mechanism to learn adaptively the importance of prior information, improving the flexibility and robustness of prior terms, and avoiding limitations similar to hand-crafted image prior terms. Consequently, we design a novel multi-scale generalized shrinkage threshold network (MGSTNet) that focuses specifically on learning deep geometric prior features to enhance image restoration. Experimental results on real and synthetic remote sensing image datasets demonstrate the superiority of our MGSTNet framework compared to existing deblurring methods.

\end{abstract}

\section{Introduction}
\IEEEPARstart{R}{ecently}, the rapid progress of remote sensing technology has led to the use of remote sensing images in various applications, such as object detection \cite{cheng2016learning, deng2018multi, yu2021new}, scene segmentation \cite{wang2022unetformer, chen2021scene, wang2020hr}, and image fusion of remote sensing \cite{wang2023mct, xie2020mhf, shen2022coupling}. However, the quality of these images is often compromised due to limitations of the imaging hardware and the complex imaging environment. As a result, the performance of algorithms in different fields is affected. To address this problem, the most cost-effective solution is to {utilize} image deblurring techniques. These techniques can improve the quality of the images, providing more distinct details compared to their low-quality images, without the need for equipment upgrades. As a result, image deblurring has become
a popular research topic in the fields of remote sensing and computer vision.

The classical image degradation model can be expressed mathematically as follow:
\begin{align}
\mathbf{g}=\mathbf{h} \ast \mathbf{u}+\mathbf{n},
\label{eq1}
\end{align}
where $\mathbf{g}$ is the degraded observational image, $\mathbf{u}$ is the latent image with sharp details, $\mathbf{h}$ is the blur kernel, $\mathbf{n}$ is the additive white Gaussian noise (AWGN), and "$*$" denotes the two-dimensional convolution operator.

Image blind deblurring is a challenging task, as it involves restoring the sharp image $\mathbf{u}$ and the blur kernel $\mathbf{h}$ from the degraded image $\mathbf{g}$. This problem is considered ill-posed because there are multiple blur kernel/high-quality image pairs that can satisfy equation (\ref{eq1}), leading to a lack of uniqueness in the solution. To address this issue, researchers have explored different approaches for image deblurring, which can be categorized into traditional model-based methods and deep learning-based methods.

Model-based methods typically involve the selection or construction of geometric prior terms as a primary means of addressing the deblurring problem. Researchers have developed a variety of geometric priors, such as self-similar priors \cite{buades2005non,dabov2007image}, total variational prior (TV) \cite{chan1998total}, gradient priors \cite{shan2008high,xu2013unnatural}, sparse coding priors \cite{luo2015removing,aharon2006k,mairal2007sparse}, hyper-Laplacian prior \cite{krishnan2009fast}, and dark channel prior \cite{pan2016blind}. These priors are used to determine the solution space for latent sharp images and blur kernels.

Despite the success of traditional model-based image deblurring techniques, there are still some issues that need to be addressed. The priors used in these methods are typically designed manually, which can lead to an imprecise representation of both the sharp image and the blurred kernel. Furthermore, it can be challenging to select the appropriate parameters for these algorithms, and the computational complexity of solving the models is quite high, resulting in a longer algorithm execution time.


In recent years, many deep learning-based approaches have been proposed for image restoration tasks, inspired by the success of deep learning in various vision tasks \cite{li2022symmetrical,li2023rgb,li2022dual,bai2023making}. These methods mainly focus on utilizing advanced deep learning techniques such as residual dense blocks \cite{zhang2020residual,zhang2018residual}, multi-scale structures \cite{zamir2021multi,cho2021rethinking,zamir2022learning}, channel attention mechanisms \cite{zamir2020learning,zamir2021multi}, spatial attention mechanisms \cite{zamir2020learning}, transformer mechanisms \cite{zamir2022restormer,wang2022uformer}, generative adversarial networks \cite{kupyn2019deblurgan,zhang2018generative,kupyn2018deblurgan}. The main goal of these methods is to learn the mapping relationship between degraded observation images and sharp images.

Many studies have demonstrated the effectiveness of deep learning techniques in image deblurring. However, there are still several challenges that need to be overcome \cite{kang2022multilayer,mou2022deep,wang2022deep}. The end-to-end  black-box design of deep learning networks makes it difficult to understand the role of the network architecture. Therefore, more transparency and interpretability are needed in these networks. In addition, many deep learning methods rely on expanding modules in terms of their depth and breadth to improve performance. This raises the question of how to design the network framework in a more rational manner.

To address the above issues, we have developed a novel approach called MGSTNet, which is a model-driven deep unfolding framework designed specifically for remote sensing image blind deblurring. This network is built on the mathematical optimization scheme of the blur kernel and the image alternating proximal gradient iteration.
The blur kernel and the latent sharp image are initially learned through a linear reconstruction step, then a deep learning module is followed to capture the deep prior features of the image/blur kernel. Furthermore, a multi-scale generalized shrinkage thresholding technique is proposed to restore lost texture details and improve the restoration of finer image details. The main contributions of this work are summarized as follows:
\begin{itemize}
\item[(1)] A MAP-based multi-scale generalized shrinkage threshold network to solve the blind deblurring problem of images is proposed, leading to an end-to-end trainable and also interpretable framework (MGSTNet).

\item[(2)] We design a kernel proximal mapping module that can be learned in the blur kernel reconstruction step. This module is able to learn and filter deep prior information of the blur kernel, resulting in a significant improvement in the evaluation of the blur kernel.

\item[(3)] We create an image deep proximal mapping module that combines a generalized shrinkage threshold operator and a multi-scale prior feature extraction block. Additionally, we incorporate an attention mechanism to adaptively learn the importance of the prior information, thus avoiding the disadvantages similar to the hand-crafted image prior terms.

\item[(4)] The experiments demonstrate that the proposed MGSTNet framework yields superior image restoration quality for blurred images with varying levels of deterioration.
\end{itemize}

The remainder of this paper is organized as follows. Section II introduces the relevant work. Section III introduces the proposed MAP-based multi-scale generalized shrinkage threshold framework for image blind deblurring in remote
sensing, and the experimental results are analyzed in Section IV. Conclusions are presented in Section V.

\begin{figure*}[htbp]
\centering
\includegraphics[width=7in]{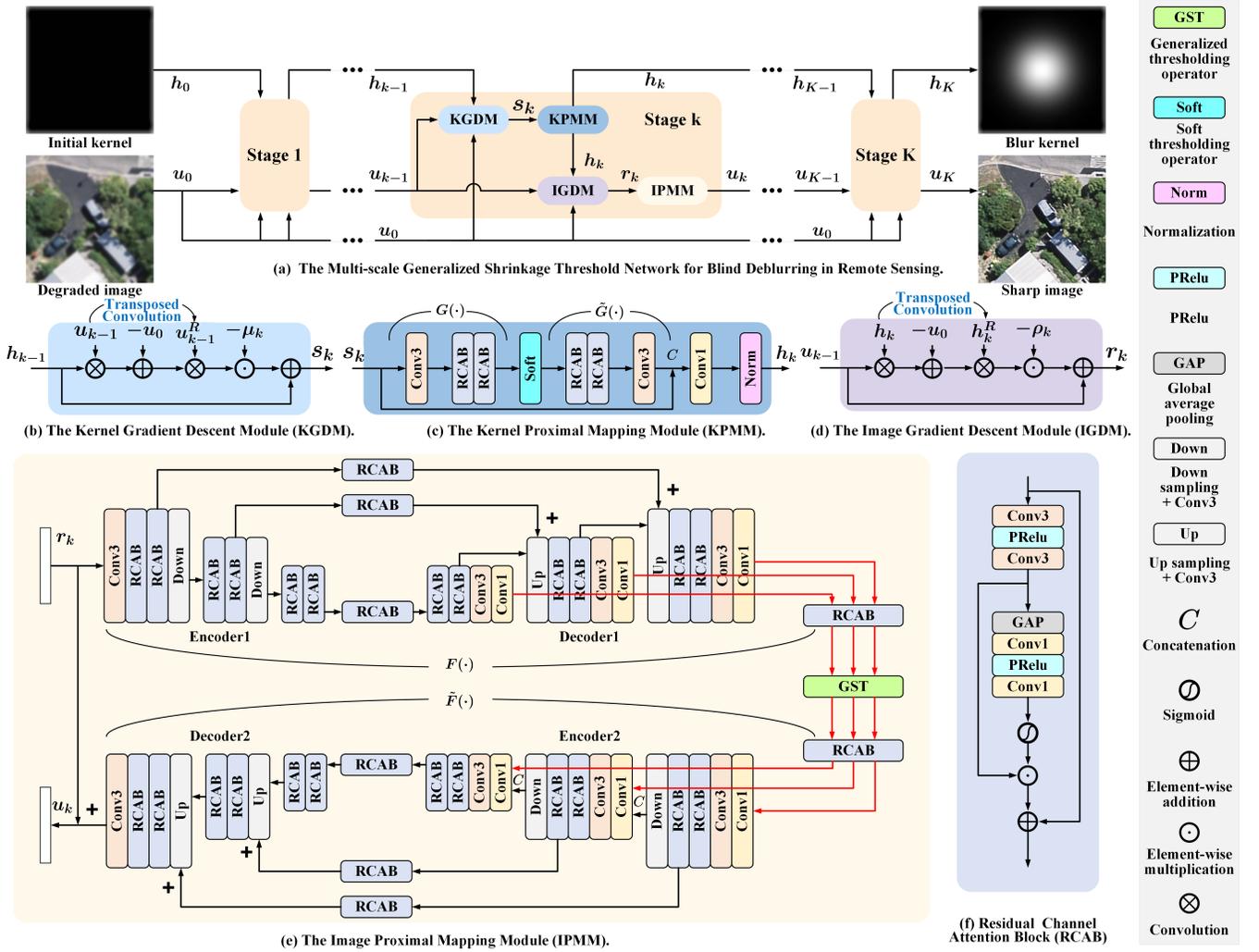}
\caption{The overall framework of the proposed MGSTNet that learns enriched feature representations for image blind deblurring. MGSTNet is based on an iterative design with alternating image/blur kernel optimizations, and its main idea is to learn the representation mapping between the image space (or kernel space) and the feature space by using an encoder-decoder module and a shrinkage threshold module.}
\label{fig1}
\end{figure*}

\section{Relevant Work}
In this section, we briefly review the relevant work on traditional model-based methods, data-driven deep learning-based methods, and model-driven deep unfolding networks.

\subsection{{Traditional Model-based Methods}}
Traditional model-based image blind deblurring methods usually treat deblurring as an ill-posed problem and solve it from a Bayesian perspective. They derive the minimum optimization problem through the maximum a posteriori (MAP) probability framework, which includes the data fidelity terms corresponding to the likelihood and the regularization terms corresponding to the image priors and blur kernel priors. To effectively constrain the solution space, researchers have done a lot of research on exploring complex image priors and blur kernel priors. For example, Chan et al. \cite{chan1998total} designed an alternating iterative TV regularization blind deblurring method to restore image edge information. Krishnan et al. \cite{krishnan2009fast} proposed a fast image deconvolution method based on a hyper-Laplacian prior. Krishnan et al. \cite{krishnan2011blind} also proposed a new regularization of the normalized sparsity measure for image blind deblurring. Zuo et al. \cite{zuo2013generalized} designed a generalized iterative shrinkage method for non-convex sparse coding for image restoration problems. Pan et al.\cite{pan2016blind} proposed an image blind deblurring algorithm based on a dark channel prior and semiquadratic splitting strategy. The above algorithms use hand-crafted regularization terms, which are difficult to handle in complex practical applications.

\subsection{Data-Driven Deep Learning-based Methods}
Deep learning-based image deblurring techniques typically employ an end-to-end network architecture to learn the non-linear mapping between blur image/sharp image pairs \cite{mao2021deep, ren2020neural, li2022supervised}. Nah et al. \cite{Nah_2017_CVPR} first introduced DeepDeblur, a network that can directly restore sharp images from blurred images in an end-to-end framework. Deepdeblur gradually restores sharp images from coarse to fine through a cascade of sub-networks. Kupyn et al. \cite{kupyn2018deblurgan, kupyn2019deblurgan} presented image deblurring in the form of generative adversarial networks. Zhang et al. \cite{zhang2020residual} proposed the residual dense block (RDB) and the residual dense network (RDN) for image restoration. Mao et al. \cite{mao2021intriguing} suggested a residual module for the frequency domain that extracts information from the frequency domain to achieve a balance between high-frequency and low-frequency information. Cui et al. \cite{cui2022selective} presented an image restoration framework denoted as a selective frequency network (SFNet), which is built on a frequency selection mechanism. Cho et al. \cite{cho2021rethinking} revisited the coarse-to-fine strategy and presented a multi-input and multi-output U-Net (MIMO-UNet) for single-image deblurring.

\subsection{Model-Driven Deep Unfolding Networks}
Recently, some deep learning-based methods have incorporated deep networks into classical model-based optimization algorithms and proposed end-to-end deep unfolding networks with optimized parameters \cite{quan2021gaussian,xu2021deep,li2020efficient,zhang2018ista, you2021ista,Fan2021,HZhu2022,Fan2023,AXLiu2023}. Quan et al. \cite{quan2021gaussian} proposed a network of deblurring that unfolds based on a Gaussian kernel mixture model and a fixed-point iteration framework. Xu et al. \cite{xu2021deep} proposed a new paradigm that combined deep unfolding and panchromatic sharpening observation models. Li et al. \cite{li2020efficient} proposed a deep unfolding method for blind deblurring based on a generalized TV regularization algorithm. Zhang and You \cite{zhang2018ista, you2021ista} proposed ISTA-Net, ISTA-Net+ and ISTA-Net++ for image compressive sensing (CS) reconstruction based on the Iterative Shrinkage-Thresholding Algorithm (ISTA). Fan et al. \cite{XHFan2023-TCI} proposed a deep geometric incremental learning framework based on the second Nesterov proximal gradient optimization. Inspired by the physical principles of the heat diffusion equations, Cui et al.\cite{ZXCui2023a} developed an interpretable diffusion learning MRI reconstruction framework that combines k-space interpolation techniques and image domain methods.

\subsection{Diffusion Models}
Diffusion models are probabilistic generative models, which have shown excellent performance in various image restoration tasks. Denoising diffusion probability models (DDPM) \cite{ho2020denoising} generate high-quality images from pure Gaussian noise through a stochastic iterative denoising process. Saharia et al. \cite{saharia2022image} presented a super resolution through repeated refinement (SR3), adapted DDPM and used the stochastic and iterative denoising process to super resolution. Xia et al.  \cite{xia2023diffir} proposed an efficient diffusion model for image restoration (DiffIR), it used a diffusion model to generate a priori representation for image restoration. Yan et al.  \cite{yan2023towards} used the conditional diffusion model to achieve high-quality dynamic range image deghosting. Chen et al.  \cite{chen2023hierarchical} proposed the hierarchical integration diffusion model (HI-Diff) for realistic image deblurring.  Cao et al. \cite{cao2023diffusion} developed a novel supervised diffusion model with two conditional modulation modules to sharpen multispectral and hyperspectral images.

Diffusion models are widely appreciated for the quality and diversity of the generated samples. However, the multistep forward/backward process of DDPM produces intermediate time-dependent variables and breaks the one-step preconditioned framework \cite{cao2023diffusion}. Consequently, it is necessary to execute multiple steps during inference to generate only one sample. This leads to computational challenges, including slow speeds due to the large number of steps required for sampling. Despite the significant amount of research conducted in this area, GANs still produce images faster \cite{FCroitoru2023Diffusion}.

As mentioned previously, traditional model-driven approaches have limitations due to the need for manual configuration of model parameter settings, making them inflexible and less robust. Consequently, applying these methods to complex real-world images becomes challenging. On the other hand, data-driven networks lack transparency and interpretability, which restricts their practical applications. To address these issues, we propose an interpretable framework that employs the iterative proximal gradient algorithm for both the blur kernel and the image. Our proposed framework incorporates an image proximal mapping module that combines multi-scale prior learning and generalized threshold operators. This module significantly improves the reconstruction results by generating high-quality image geometry priors.

\section{Methodology}
\subsection{The MAP Framework}
The MAP method is a powerful tool that is commonly used to solve inverse problems. It is based on Bayesian inference and offers a systematic approach to regularizing solutions based on a set of observations. This method is particularly valuable for addressing ill-posed problems in image processing. To provide a better understanding of our work, we now give a brief overview of the MAP framework that is employed in blind deblurring methods \cite{ren2020neural,li2022supervised}. These methods aim to estimate both the image $\mathbf{u}$ and the blur kernel $\mathbf{h}$ from a given blurred image $\mathbf{g}$. The problem can be defined as follows:
\begin{equation}
\begin{split}
(\mathbf{u}, \mathbf{h}) &=\underset{\mathbf{u},\mathbf{h}}{\operatorname{argmax}} \enspace p(\mathbf{h}, \mathbf{u} \mid \mathbf{g})\\
&=\underset{\mathbf{u},\mathbf{h}}{\operatorname{argmax}} \enspace p(\mathbf{g} \mid \mathbf{h}, \mathbf{u}) p_{u}(\mathbf{u}) p_{h}(\mathbf{h}),
\end{split}
\label{eq3}
\end{equation}
where $p(\mathbf{h}, \mathbf{u}\mid\mathbf{g})$ is the posterior probability, $p(\mathbf{g}\mid\mathbf{h},\mathbf{u})$ represents the likelihood function. The $p_{u}(\mathbf{u})$ and $p_{h}(\mathbf{h})$ denote the probability density functions (statistical priors) of $\mathbf{u}$ and $\mathbf{h}$, respectively.

According to \eqref{eq1} and the general formulation of MAP-based blind deblurring, we define the likelihood probability term as
\begin{equation}
p(\mathbf{g} \mid \mathbf{h}, \mathbf{u}) \propto e^{-\cfrac{\|\mathbf{h}\ast \mathbf{u}-\mathbf{g}\|_{2}^{2}}{2 \sigma_1^{2}}}
\label{eq5}
\end{equation}
and further define the two prior probability terms as follows
\begin{equation}
\begin{split}
p_{u}(\mathbf{u}) \propto e^{-\cfrac{\|\Psi(\mathbf{u})\|_{p}^{p}}{2 \sigma_2^{2}}},\quad
p_{h}(\mathbf{h}) \propto e^{-\cfrac{\|\Phi(\mathbf{h})\|_{1}^{1}}{2 \sigma_3^{2}}}.
\end{split}
\label{eq7}
\end{equation}

Taking the negative logarithm of both sides of \eqref{eq3} yields a MAP estimator, which is expressed as follows:
\begin{equation}
\underset{\mathbf{u},\mathbf{h}}{\operatorname{argmin}} -\log p(\mathbf{g} \mid \mathbf{h}, \mathbf{u})-\log p_{u}(\mathbf{u})-\log p_{h}(\mathbf{h}),\label{eq4}
\end{equation}
{where $\log p_{u}(\mathbf{u})$ and $\log p_{h}(\mathbf{h})$} are the logarithms of the prior distributions of $\mathbf{u}$ and $\mathbf{h}$, respectively. Then, \eqref{eq4} can be formulated as a regularized optimization as follows:
\begin{equation}
\begin{split}
\underset{\mathbf{u},\mathbf{h}, \|\mathbf{h}\|_{1}=1}{\operatorname{argmin}} &\; \|\mathbf{h} \ast \mathbf{u}-\mathbf{g}\|_{2}^{2}+\lambda_1\|{\Psi} ({\mathbf{u}})\|_{p}^{p}+\lambda_2\|{\Phi} ({\mathbf{h}})\|_{1}^{1},
\end{split}
\label{eq8}
\end{equation}
where $\|\mathbf{h} \ast \mathbf{u}-\mathbf{g}\|_{2}^{2}$ represents the data fidelity term, the $\|{\Psi} ({\mathbf{u}})\|_{p}^{p}$ and $\|{\Phi} ({\mathbf{h}})\|_{1}^{1}$ are the regularization terms induced by the priors imposed on image and kernel, {$\lambda_1=\sigma_1^{2}/\sigma_2^{2}$ and $\lambda_2=\sigma_1^{2}/\sigma_3^{2}$ are the regularization coefficients.} $\Phi(\cdot)$ and $\Psi(\cdot)$ are transformations from the image/blur kernel spaces to the feature space, such as the Fourier transform, the discrete cosine transform, or the discrete wavelet transform.

In traditional methods and machine learning methods, the proximal gradient method is typically used to solve (\ref{eq8}). It can efficiently solve compound optimization problems with $L_{1}$ or $L_{p}$ non-smooth regularization terms by introducing the soft-threshold operator and the generalized threshold operator. However, there are still some limitations. (1) For complicated nonlinear transformations $\Phi(\cdot)$ and $\Psi(\cdot)$, the traditional proximal gradient method often takes numerous iterations to produce satisfying results, which demands a substantial amount of computing. (2) These hand-crafted priors may not be able to accurately capture the feature-complex and geometrically diverse nature of the sharp image and the blur kernel, which can lead to inaccurate or unnatural deblurring results. To address these, we introduce a deep prior learning framework. Using a large amount of data, this framework can learn complex prior information, thus enhancing the generalizability of image deblurring algorithms.

\subsection{Proposed Deblurring Framework}
In the previous section, we introduced a MAP framework and obtained a minimization problem (\ref{eq8}). To address the problem of multiple coupled variables and the geometrical diverse nature of the solutions, we propose the use of a learnable alternating iteration scheme, where transformations ${\Psi}(\cdot)$ and ${\Phi}(\cdot)$, which map the image/blur kernel spaces to the feature space, are replaced by learnable non-linear transformation modules $F(\cdot)$ and $G(\cdot)$. The goal is to utilize two deep prior modules to learn the knowledge of geometric prior features from a large dataset of images and to fit prior non-linear transformations. This allows us to split the problem (\ref{eq8}) into two subproblems: one involving the latent sharp image $\mathbf{u}$ and the other involving the blur kernel $\mathbf{h}$, which can be expressed as follows:
\begin{equation}
\underset{\mathbf{h},\; \| \mathbf{h}\| _{1}=1}{\operatorname{argmin}}\; \| \mathbf{h} \ast \mathbf{u}-\mathbf{g}\| _{2}^{2}+2\lambda_1\| {G} ({\mathbf{h}})\|_{1}^{1}, \label{eq10} \end{equation}
and
\begin{equation} \underset{\mathbf{u}}{\operatorname{argmin}}\; \| \mathbf{h} \ast \mathbf{u}-\mathbf{g}\| _{2}^{2}+2\lambda_2\| {F} ({\mathbf{u}})\|_{p}^{p}. \label{eq11} \end{equation}

Using the learnable framework described above, we designed a novel image blind deblurring network architecture, as illustrated in Fig. \ref{fig1} (a), which consists of a series of $K$ reconstruction stages. Each stage includes both the blur kernel reconstruction and image reconstruction.

\subsubsection{Blur Kernel Reconstruction}
Mathematically, the classical optimization solving problem \eqref{eq10} leads to two-step updates, which involve the gradient descent iteration $f_{\text{Ker},1}(\cdot)$ of the blur kernel and the proximal mapping step $f_{\text{Ker},2}(\cdot)$ of the blur kernel.

First, the linear reconstruction of the blur kernel $\mathbf{s}_{k}$ can be obtained through gradient descent, that is,
\begin{equation}
\begin{split}
\mathbf{s}_{k} &= f_{\text{Ker},1}(\mathbf{h}_{k-1},\mathbf{u}_{k-1},\mathbf{g})\\
               &= \mathbf{h}_{k-1}-\mu_{k} \mathbf{u}_{k-1}(-x,-y)\ast(\mathbf{u}_{k-1} \ast \mathbf{h}_{k-1}-\mathbf{g}),
\end{split}
\label{eq13}
\end{equation}
where $\mu_{k}$ is the step size in the $k$-step iteration. Then, the kernel proximal-point mapping step $f_{\text{Ker},2}(\cdot)$ can be derived as follows:
\begin{equation}
\begin{split}
\mathbf{h}_{k-\frac{1}{2}} &=f_{\text{Ker},2}(\mathbf{s}_{k})\\
&= \textbf{prox}_{\lambda_{1,k},\|G(\cdot)\|_{1}^{1}}(\mathbf{s}_{k})\\
&= \underset{\mathbf{h}}{\operatorname{argmin}}\; \|\mathbf{h}-\mathbf{s}_{k}\|_{2}^{2}+2\lambda_{1,k} \|G(\mathbf{h})\|_{1}^{1}.
\end{split}
\label{eq15}
\end{equation}

\textbf{Kernel Gradient Descent Module}.
Inspired by ISTA-Net+ et al. \cite{zhang2018ista,mou2022deep,you2021ista,Fan2021,Fan2023}, we employ a kernel gradient descent module (KGDM) $f_{\text{KGDM}}(\cdot)$ with a learnable parameter $\mu_{k}$ to evolve the linear reconstruction gradient flow $f_{\text{Ker},1}(\cdot)$ in (\ref{eq13}), that is,
\begin{equation}
\begin{split}
\mathbf{s}_{k} &= f_{\text{KGDM}}(\mathbf{h}_{k-1},\mathbf{u}_{k-1},\mathbf{g},\mu_k)\\
&= \mathbf{h}_{k-1}-\mu_{k} \mathbf{u}_{k-1}(-x,-y)\ast(\mathbf{u}_{k-1} \ast \mathbf{h}_{k-1}-\mathbf{g}).
\end{split}
\label{eq13b}
\end{equation}

\textbf{Kernel Proximal Mapping Module}.
Additionally, we propose an effective kernel proximal mapping module (KPMM) $f_{\text{KPMM}}(\cdot)$ to reconstruct more image features by learning the proximal-point step $f_{\text{Ker},2}(\cdot)$ of the kernel $\mathbf{h}$ in \eqref{eq15}, that is,
\begin{equation}
\mathbf{h}_{k-\frac{1}{2}} =f_{\text{KPMM}}(\mathbf{s}_{k})= \mathbf{s}_{k} + \tilde{G}_{k}(\text{Soft}(G_{k}(\mathbf{s}_{k}),\theta_{1,k})),
\label{eq18}
\end{equation}
where $\theta_{1,k}$ is a threshold that is linearly related to $\lambda_{1,k}$ in (\ref{eq15}).
The soft-thresholding operator $\text{Soft}(\cdot)$ in (\ref{eq18}) is commonly used to solve the optimization problems with a $\ell_1$-norm regularization term, which can be expressed as:
\begin{align}
{\text{Soft}(y,\theta)} =
\begin{cases} 0,&{\text{if}}\  {y }\leq\theta, \\ {\text{sgn}(y)(y -\theta),}&{\text{if}}\  {y }>\theta, \
\end{cases}
\label{eq19}
\end{align}
where $\theta$ is the threshold value. For the $k$ stage with an architecture of unshared parameters, the KPMM module in (\ref{eq18}) is composed of a module encoded with deep kernel prior feature extraction $G_k(\cdot)$, a {soft-thresholding} operator $\text{Soft}(\cdot)$ with a learnable threshold $\theta_{1,k}$, a module decoded with deep kernel prior feature extraction $\tilde{G}_k(\cdot)$ (inverse transformation) and a long skip connection, as illustrated in Fig. \ref{fig1}(c).

\textbf{Deep Kernel Prior Extractors $G_k(\cdot)$ and $\tilde{G}_k(\cdot)$}.
Fig. \ref{fig1}(c) illustrates the architectures of our deep kernel prior feature extractor $G_k(\cdot)$ and its inverse transformation $\tilde{G}_k(\cdot)$. {To extract the shallow prior feature of the blur kernel, a $3\times3$ convolution is used. Additionally, two residual channel attention blocks (RCAB)\cite{zhang2018image} are used to extract deep kernel prior features and balance the prior features of each channel (\emph{see} Fig. \ref{fig1} (f) for RCAB).} For the inverse transformation $\tilde{G}_k(\cdot)$, a symmetric structure is employed to map the prior features back to the blur kernel space.

Using KPMM with a deep kernel prior, a high-quality blur kernel can be reconstructed. To ensure that the kernel $\mathbf{h}\geq 0$ meets the commonly used physical constraint $\|\mathbf{h}\|_1=1$, a normalization layer is added to the end of the blur kernel reconstruction, resulting in
\begin{align}
\mathbf{h}_{k}&= \mathbf{h}_{k-\frac{1}{2}}/||\mathbf{h}_{k-\frac{1}{2}}||_{1}.\label{eq20}
\end{align}

\subsubsection{Image Deblurring Reconstruction}
It is difficult to recover a high-quality image $\mathbf{u}$ with only low-frequency information due to the low-pass nature of the blur kernel {$\mathbf{h}_{k}$}. Consequently, the geometric priors of the image must be learned during the image deblurring stage. This stage consists of an {image gradient descent linear reconstruction step and an image proximal mapping step}, thus the solutions can be expressed as $ f_{\text{Img},1}(\cdot)$ and $f_{\text{Img},2}(\cdot)$, respectively,
\begin{equation}
\begin{split}
\mathbf{r}_{k} &= f_{\text{Img},1}(\mathbf{u}_{k-1}, \mathbf{h}_{k}, \mathbf{g})
\\
&= \mathbf{u}_{k-1} - \rho_{k} \mathbf{h}_{k}(-x,-y)\ast(\mathbf{h}_{k} \ast \mathbf{u}_{k-1}-\mathbf{g})
\end{split}
\label{eq22}
\end{equation}
and
\begin{equation}
\begin{split}
\mathbf{u}_{k} &=f_{\text{Img},2}(\cdot)= \textbf{prox}_{\lambda_{2,k},\|{F_{k}}(\cdot)\|_{p}^{p}}(\mathbf{r}_{k})
\\
&= \underset{\mathbf{u}}{\operatorname{argmin}}\|\mathbf{u}-\mathbf{r}_{k}\|_{2}^{2}+2\lambda_{2,k} \|{F_{k}}(\mathbf{u})\|_{p}^{p},
\end{split}
\label{eq24}
\end{equation}
where $\rho_{k}$ is the step-length.

\textbf{Image Gradient Descent Module}. In the proposed framework, we employ an image gradient descent module (IGDM) $f_{\text{IGDM}}(\cdot)$ with a learnable parameter $\rho_{k}$ to evolve the linear reconstruction gradient flow $f_{\text{Img},1}(\cdot)$ in (\ref{eq22}), that is,
\begin{equation}
\begin{split}
\mathbf{r}_{k} &= f_{\text{IGDM}}(\mathbf{u}_{k-1}, \mathbf{h}_{k}, \mathbf{g},\rho_k)
\\
&= \mathbf{u}_{k-1} - \rho_{k} \mathbf{h}_{k}(-x,-y)\ast(\mathbf{h}_{k} \ast \mathbf{u}_{k-1}-\mathbf{g}).
\end{split}
\label{eq22b}
\end{equation}

\textbf{Image Proximal Mapping Module}. Additionally, we design an image proximal-point mapping module (IPMM) $f_{\text{IPMM}}(\cdot)$ to learn the image proximal mapping solution $f_{\text{Img},2}(\cdot)$ in (\ref{eq24}), which is expressed as follows:
\begin{equation}
\begin{split}
\mathbf{u}_{k} &=f_{\text{IPMM}}(\cdot)\\
&= \mathbf{r}_{k} + \tilde{F}_{k}(\text{GST}_{p}(F_{k}(\mathbf{r}_{k}),\theta_{2,k})).
\end{split}
\label{eq26}
\end{equation}

The proposed IPMM module consists of a deep image prior block $F(\cdot)$, a generalized thresholding operator $\text{GST}_{p}(\cdot)$, an inverse deep image prior block $\tilde{F}(\cdot)$, and a long skip connection, as illustrated in Fig. \ref{fig1}(e). The generalized thresholding operator \cite{zuo2013generalized} is formulated as
\begin{align}
\text{GST}_{p}(y,\theta) = \begin{cases}
0,&{\text{if}}\ {|y| }\leq \tau_{p}(\theta), \\
{\text{sgn}(y)S_{p}(|y| ; \theta),}&{\text{if}}\ {|y| }> \tau_{p}(\theta),
\end{cases}
\label{eq27}
\end{align}
where $\tau_{p}(\theta)= (2\theta(1-p))^{\frac{1}{2-p}} + \theta p(2\theta(1-p))^{\frac{p-1}{2-p}}$ is the threshold determined by the norm coefficient $p$ and the $\theta$, and $\theta$ is linearly related to the regularization coefficient $\lambda_{2,k}$ in (\ref{eq24}) (also \emph{see} \cite{zhang2018ista}). The norm coefficient $p$ is mapped to a value between 0 and 1 using a sigmoid function, that is, $p = \text{Sigmoid}(p_{0})$. The generalized thresholding operator has the same thresholding rule as the soft-thresholding operator. In the shrinkage rule, the generalized thresholding operator assigns $\text{sgn}(y)S_{p}(|y|;\theta)$ to $\text{GST}_{p}(y,\theta)$. The $S_{p}(|y|;\theta)$ is obtained by iterating the following formula in a fixed point iteration \cite{zuo2013generalized} as follows:
\begin{align}
S_{p,n}(|y| ; \theta) = |y| - \theta p(S_{p,n-1}(|y| ; \theta) + \delta )^{p-1},\label{eq29} \
\end{align}
where $\delta=10^{-5}$ is a small perturbation term to ensure the network stability, The fixed-point iteration number $n$ is set as $3$ and the initialization is $S_{p,0}(|y|;\theta)=|y|$.

\textbf{Deep Image Prior Extractors $F_{k}(\cdot)$ and $\tilde{F}_{k}(\cdot)$}.
The images obtained by remote sensing contain a wealth of information about land objects, topography, and a variety of textures and structures that are more intricate than a blur kernel. To accurately capture their prior features, it is essential to take into account complex information from multiple perspectives, such as color, texture, shape, and spatial relationships. In this work, we propose the priori feature transformations (deep image prior extractors) denoted as $F_{k}(\cdot)$ and $\tilde{F}_{k}(\cdot)$ that are activated by deep learning, as illustrated in Fig.\ref{fig1}(e). These transformations must be sufficiently powerful and reliable.

Our deep image prior extractor $F_{k}(\cdot)$ is composed of an encoder and a decoder to take advantage of multi-scale feature maps, and a symmetrical structure is employed to construct $\tilde{F}_{k}(\cdot)$. The success of the encoder-decoder architecture has demonstrated that learning the information about the multi-scale image features accurately can improve the performance of the model \cite{ren2020neural,li2022supervised,tsai2022banet,zamir2021multi}. The encoder-decoder architecture can learn semantic image information by going through multiple upsampling/downsampling operations and extracting feature information at different scales to acquire a sufficiently large receptive field.

The input shallow features in the IPMM module are first extracted using the first convolutional layer of the deep image prior extractor $F_{k}(\cdot)$. Subsequently, RCABs are used to extract features at three different scales. Moreover, RCABs are also used to manage skip connections between the encoder and the decoder. The decoder generates multi-scale features, which are further modified through convolution and RCAB operations, and finally outputted as image prior features.

As a result, we utilize a generalized shrinkage threshold operator to extract the multi-scale prior features. Subsequently, we employ the second encoder-decoder $\tilde{F}(\cdot)$ to transform these features back into the image space. This process is illustrated in Fig. \ref{fig1}(e), where RCAB and convolution are utilized to adjust and combine the multi-scale prior features with the encoder features. In contrast to MIMO-UNet+, which generates multi-scale outputs within the image space and utilizes multi-scale loss terms for network training optimization, our proposed framework generates a nonlinear representation from the image space to the prior feature space.

Previous algorithms \cite{zhang2018ista, you2021ista} have mainly focused on single-scale applications while learning prior features. In contrast, our framework incorporates multi-scale shrinkage threshold operators, {as illustrated in Fig. \ref{fig1}(e). This approach enables the network to learn the prior} features for each scale separately. To achieve a more comprehensive and diverse prior representation, a decoder is employed to produce multi-scale outputs and map the image into the multi-scale feature space.

Finally, we design a MAP-based multi-scale generalized shrinkage threshold framework to solve the image blind deblurring problem that is guided by the proximal gradient descent iterative algorithm for two subproblems. Our proposed framework incorporates deep prior module and multi-scale prior feature extraction blocks, which have stronger expressive power than traditional model-based algorithms. Additionally, each unfolding module in our approach corresponds to an iterative step of the optimization algorithm, resulting in a model that is easy to interpret and can effectively learn representations. Through a theoretical framework, our method demonstrates improved robustness when evaluating complex degraded images. Furthermore, our approach produces more reliable results compared to data-driven networks.

\subsection{Loss Function}
The MGSTNet framework is comprised of two components in each stage: image deblurring reconstruction and blur kernel reconstruction. To optimize the parameters of the two reconstruction parts, we minimize the total loss, represented as
\begin{equation}
\begin{gathered}
\mathcal{L}_{Total}=\mathcal{L}_{Char} + \alpha \mathcal{L}_{K},
\end{gathered}
\label{eq30}
\end{equation}
which consists of two components: the Charbonnier loss $\mathcal{L}_{Char}$ and the kernel loss $\mathcal{L}_{K}$. The weight parameter $\alpha$ is set to 0.05. The $\mathcal{L}_{Char}$ is employed to guide the optimization direction of the image proximal mapping module and determine the value of the learnable parameter $\rho_{k}$ in the image gradient descent module. On the other hand, $\mathcal{L}_{K}$ is used to guide the optimization direction of the kernel proximal mapping module and select the value of the learnable parameter $\mu_{k}$ in the kernel gradient descent module.

\textbf{(1) Charbonnier loss}.
To achieve high-quality image reconstruction, the Charbonnier loss is used at each stage to quantify the discrepancy between the reconstructed image and the ground truth image \cite{lai2018fast}. This is expressed as follows:
\begin{equation}
\mathcal{L}_{Char}= \sum_{k=1}^{K}\sqrt{\|{\mathbf{u}_{k}}-{\mathbf{u}_{gt}}\|^{2}+\varepsilon_{1}^{2}},
\label{eq31}
\end{equation}
where $\varepsilon_{1}=1\times10^{-3}$ and $k$ represents the stage index. The Charbonnier loss effectively prevents over-smoothing of the results, allowing the preservation of image edges and details while disregarding small noise and other disturbances. This ensures the robustness and performance of the proposed framework.

\textbf{(2) Kernel loss}. In the context of reconstructing the blur kernel, we employ the ground truth image and the reconstructed blur kernel at each stage to generate a blurry image. The blurry image is then used to formulate the loss function with the degraded image. Therefore, the kernel loss can be expressed as follows:
\begin{equation}
\mathcal{L}_{K}=\sum_{k=1}^{K} \left\| \mathbf{h}_{k}*\mathbf{u}_{gt}-\mathbf{g}\right\|_{1}.
\label{eq32}
\end{equation}

Using this loss function, we establish a connection between sharp and blurred images, which offers guidance for reconstructing the blur kernel. This loss function can provide accurate spatial blur supervision even in the presence of noise, ensuring reliable reconstruction of the blur kernel and further improving the performance of image deblurring.

\section{Experiments}
In this section, we assess the performance of our approach in image deblurring tasks and compare it with existing state-of-the-art methods on blurred aerial and remote sensing image datasets.

\subsection{Implementation Details}
\subsubsection{\textbf{Dataset}} In our experiments, three widely used benchmark remote sensing datasets, AIRS, AID and WHU-RS19, are utilized. The AIRS and AID datasets are used to train our proposed model, while the AIRS and WHU-RS19 datasets are employed to evaluate the deblurring performance of all approaches.

\textbf{AIRS}\cite{chen2019temporary}: The AIRS dataset consists of more than 1,000 aerial images with a spatial resolution of 0.075 m/pixel and a size of $10000\times 10000$.

\textbf{AID}\cite{xia2017aid}: The AID dataset is composed of 10,000 aerial images, each with a size of $600 \times 600$ pixels and a spatial resolution of 0.5 m/pixel. It includes 30 distinct land-use categories, such as airports, farmland, beaches, deserts, etc.

\textbf{WHU-RS19}\cite{dai2010satellite}: The WHU-RS19 dataset consists of 1005 remote sensing images, each with a size of $600\times600$ pixels and a maximum spatial resolution of 0.5 m/pixel. These images contain 19 distinct land types, including Airports, Bridges, Mountains, Rivers, and more.

\subsubsection{\textbf{Compared Methods}} To evaluate the effectiveness of our method in image deblurring, we conducted a comparison with nine other methods. These include the normalized sparsity measure method (NSM) \cite{krishnan2011blind}, the DeblurGAN-v2 method that uses generative adversarial networks for single image deblurring \cite{kupyn2019deblurgan}, the Gaussian kernel mixture network (GKMNet) for single image defocus deblurring \cite{quan2021gaussian}, the multi-input multi-output U-Net (MIMO-UNet+) \cite{cho2021rethinking}, the residual dense network (RDN) \cite{zhang2020residual}, and the selective frequency network (SFNet) \cite{cui2022selective}. All of these methods are designed for general image restoration purposes. To further validate the performance of our MGSTNet, we also compared it with three methods specifically designed for remote sensing super-resolution: the local-global combined network (LGCNet) \cite{7937881}, the mixed high-order attention network (MHAN) \cite{9151234} and the transformer-based enhancement network (TransENet) \cite{9654169}. We set the super-resolution scale factors to 1 for comparison. All the compared methods in this study were trained using the same dataset and the implementation provided by the authors on their homepage or Github, as described in their respective works.

\begin{figure}[htbp]
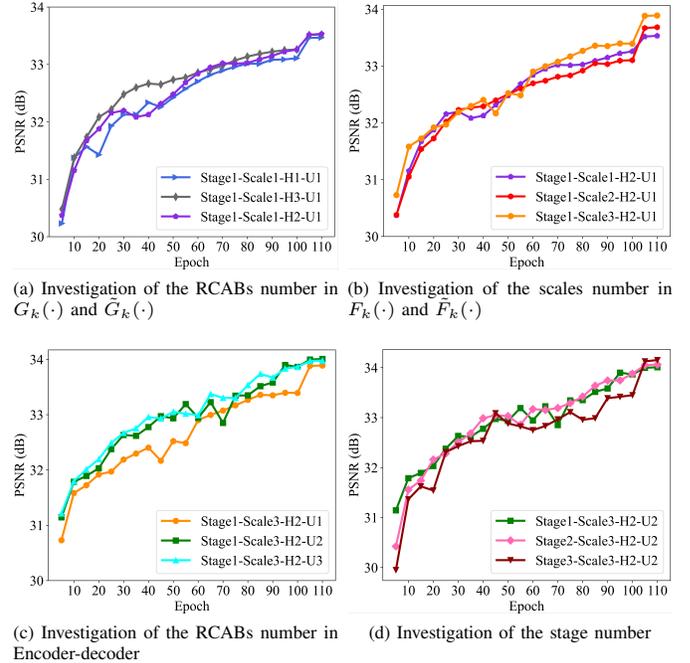

    \centering
    \subfloat[Investigation of the RCABs number in $G_{k}(\cdot)$ and $\tilde{G}_{k}(\cdot)$]{\includegraphics[width=1.7in]{figs/DL-Deblur-kernel-RCABs.eps}}%
    \label{a}
    \subfloat[Investigation of the scales number in $F_{k}(\cdot)$ and $\tilde{F}_{k}(\cdot)$]{\includegraphics[width=1.7in]{figs/DL-Deblur-SCALE.eps}}%
    \label{b}

    \subfloat[Investigation of the RCABs number in Encoder-decoder]{\includegraphics[width=1.7in]{figs/DL-Deblur-img.eps}}%
    \label{c}
    \subfloat[Investigation of the stage number]{\includegraphics[width=1.7in]{figs/DL-Deblur-STAGE.eps}}%
    \label{d}

    \caption{Convergence analysis of MGSTNet for image deblurring with different structure.}
    \label{fig2}
\end{figure}

\subsubsection{\textbf{Parameter Setting}} During the training phase, we use the AID dataset as the reference image source. To generate the degraded blur images, we crop a $256 \times 256$ patch from the sharp images and apply the equation \eqref{eq1}. The size of the isotropic Gaussian blur kernel in \eqref{eq1} is set to $15 \times 15$, and the standard deviation $\sigma$ is randomly selected from the range of [0.2, 4.0]. These image pairs are then used as inputs for training the network.

During the testing phase, we apply the same random Gaussian blur to WHU-RS19 and use the synthesized images of WHU-RS19 as a test dataset to evaluate the performance. To assess the image deblurring performance of WHU-RS19, {we utilize} peak signal-to-noise ratio (PSNR), structural similarity (SSIM) \cite{wang2004image}, Feature Similarity (FSIM) \cite{zhang2011fsim}, visual information fidelity (VIF) \cite{sheikh2006image}, and information fidelity criterion (IFC) \cite{sheikh2005information}.

In the experimental part of AIRS, we generate 1100 sharp image patches from the AIRS dataset by randomly cropping them with a size of $256 \times 256$. To create pairs of sharp and blurred images, we randomly select a standard deviation for the Gaussian kernel from the range of 1.0 to 4.0, using the equation (\ref{eq1}). In these pairs, 1000 are used for training and 100 for testing. To further assess the performance of each deblurring network in more complex degradation scenarios, we introduce Gaussian noise with a noise level of $10^{-3}$ to blurred images from the AIRS dataset, which originally have a noise level close to 0.

In numerical experiments on the AID and WHU-RS19 datasets, we train our network for 50 epochs. The learning rate is initially set to $10^{-4}$ and decreases to $10^{-5}$ at 40 epochs. For the AIRS dataset, we train our network for 110 epochs. The initial learning rate is $10^{-4}$ and decreases to $10^{-5}$ at 100 epochs. We utilize the Adam optimizer \cite{kingma2014adam} and conduct all experiments using PyTorch.

\begin{table}
    \caption{Ablation study about the number of RCABs in the modules $G(\cdot)$ and $\tilde{G}(\cdot)$ by the proposed MGSTNet on the AIRS Dataset.}
    \label{table1}
    \renewcommand\arraystretch{1.1}
    \setlength\tabcolsep{2pt}
    \centering
    \begin{tabular*}{\hsize}{@{}@{\extracolsep{\fill}}lccc@{}}
    \toprule[1.5pt]
\multicolumn{1}{l}{Method} &  \multicolumn{3}{c}{MGSTNet} \\
\midrule[0.95pt]
{RCABs number in $G(\cdot)$ and $\tilde{G}(\cdot)$}         & 1            & 2            & 3  \\
\midrule[0.95pt]
PSNR  & 33.47        & 33.53        & 33.48   \\
SSIM  & 0.8631       & 0.8642       & 0.8633  \\
FSIM  &0.9126       & 0.9133       & 0.9123  \\
VIF   & 0.5490       & 0.5510       & 0.5499  \\
IFC   & 4.8654       & 4.9008       & 4.8835  \\
\bottomrule[1.5pt]
    \end{tabular*}
\end{table}

\begin{table}
    \caption{Ablation study about the blur kernel proximal mapping module by the proposed MGSTNet framework on the AIRS Dataset.}
    \label{table1b}
    \renewcommand\arraystretch{1.1}
    \setlength\tabcolsep{2pt}
    \centering
    \begin{tabular*}{\hsize}{@{}@{\extracolsep{\fill}}lcc@{}}
    \toprule[1.5pt]
\multicolumn{1}{l}{Method}& \multicolumn{2}{c}{MGSTNet}   \\
\midrule[0.95pt]
{Kernel index}      & $s/||s||_{1}$ &$h$ \\
\midrule[0.95pt]
MNC                         &0.7036   &0.9447  \\
MSE                         & $7.9209\times10^{-3}$  & $2.0858\times10^{-5}$  \\
RMSE                        & $3.8036\times10^{-2}$  & $3.5029\times10^{-3}$  \\
\bottomrule[1.5pt]
    \end{tabular*}
\end{table}

\begin{table}
    \caption{Ablation study about the number of scales in the modules $F(\cdot)$ and $\tilde{F}(\cdot)$ by the proposed MGSTNet on the AIRS Dataset.}
    \label{table2}
    \renewcommand\arraystretch{1.1}
    \setlength\tabcolsep{2pt}
    \centering
    \begin{tabular*}{\hsize}{@{}@{\extracolsep{\fill}}lccc@{}}
    \toprule[1.5pt]
\multicolumn{1}{l}{Method} &  \multicolumn{3}{c}{MGSTNet}  \\
\midrule[0.95pt]
{Scales number in $F(\cdot)$ and $\tilde{F}(\cdot)$}         & 1                   &2              & 3     \\
\midrule[0.95pt]
PSNR & 33.53   & 33.66    & 33.88  \\
SSIM & 0.8642  & 0.8736   & 0.8832  \\
FSIM & 0.9133  & 0.9180   & 0.9247  \\
VIF  & 0.5510  & 0.5545   & 0.5702  \\
IFC  & 4.9008  & 4.9459   & 5.1085  \\
\bottomrule[1.5pt]
    \end{tabular*}
\end{table}

\begin{table}
    \caption{Ablation study about the number of RCABs in the modules $F(\cdot)$ and $\tilde{F}(\cdot)$ by the proposed MGSTNet on the AIRS Dataset.}
    \label{table3}
    \renewcommand\arraystretch{1.1}
    \setlength\tabcolsep{2pt}
    \centering
    \begin{tabular*}{\hsize}{@{}@{\extracolsep{\fill}}lccc@{}}
    \toprule[1.5pt]
\multicolumn{1}{l}{Method} &  \multicolumn{3}{c}{MGSTNet}  \\
\midrule[0.95pt]
{RCABs number in $F(\cdot)$ and $\tilde{F}(\cdot)$}         & 1                   &2              & 3     \\
\midrule[0.95pt]
PSNR & 33.88  & 33.99   & 33.97   \\
SSIM & 0.8832 & 0.8856  & 0.8856  \\
FSIM & 0.9247 & 0.9258  & 0.9263  \\
VIF & 0.5702 & 0.5741 & 0.5756    \\
IFC & 5.1085 & 5.1762 & 5.2082    \\
\bottomrule[1.5pt]
    \end{tabular*}
\end{table}

\begin{table}
    \caption{Ablation study about the stage number of MGSTNet on the AIRS Dataset.}
    \label{table4}
    \renewcommand\arraystretch{1.1}
    \setlength\tabcolsep{2pt}
    \centering
    \begin{tabular*}{\hsize}{@{}@{\extracolsep{\fill}}lccc@{}}
    \toprule[1.5pt]
\multicolumn{1}{l}{Method} & \multicolumn{3}{c}{MGSTNet}  \\
\midrule[0.95pt]
{Stage number}         & 1                   &2              & 3     \\
\midrule[0.95pt]
PSNR      & 33.99   & 34.04  & 34.13   \\
SSIM      & 0.8856  & 0.8878 & 0.8881  \\
FSIM      & 0.9258 & 0.9271  & 0.9267   \\
VIF & 0.5741 & 0.5772 & 0.5761   \\
IFC & 5.1762 & 5.2479 & 5.2205   \\
\bottomrule[1.5pt]
    \end{tabular*}
\end{table}

\begin{table*}
    \caption{{Comparison of MGSTNet with Loss weight parameter $\alpha$ values on AIRS dataset.}}
    \label{table7}
    \renewcommand\arraystretch{1.1}
    \setlength\tabcolsep{2pt}
    \centering
    \begin{tabular*}{\hsize}{@{}@{\extracolsep{\fill}}lcccccccccccccc@{}}
    \toprule[1.5pt]
 {\multirow{3}{*}{Loss weight $\alpha$}}  & \multicolumn{7}{c}{{AIRS}} & \multicolumn{7}{c}{{AIRS(+)}} \\
 \cmidrule(r){2-8} \cmidrule(r){9-15}
 & \multicolumn{2}{c}{{Kernel Index}} & \multicolumn{5}{c}{{Image Index}} & \multicolumn{2}{c}{{Kernel Index}} & \multicolumn{5}{c}{{Image Index}}\\
 \cmidrule(r){2-3} \cmidrule(r){4-8}  \cmidrule(r){9-10}\cmidrule(r){11-15}
&  {MNC} &  {MSE} &  {PSNR} &  {SSIM} &  {FSIM} &  {VIF} &  {IFC}  &  {MNC} &  {MSE}   &  {PSNR}&  {SSIM} &  {FSIM} &  {VIF} &  {IFC} \\
\midrule[0.95pt]
 {0}              &  {0.9426}            &  {$1.99\times10^{-5}$}      &  {34.03}           &  {0.8859}          &  {\textbf{0.9267}} &  {0.5737}              &  {5.2089}
                            &  {0.8853}            &  {$3.10\times10^{-5}$}      &  {33.74}           &  {0.8785}          &  {\textbf{0.9224}} &  {0.5582}              &  {5.0104} \\
 {0.001}          &  {0.8226}            &  {$5.11\times10^{-5}$}      &  {33.83}           &  {0.8851}          &  {0.9236}          &  {0.5696}              &  {5.0588}
                            &  {0.9396}            &  {$1.94\times10^{-5}$}      &  {\textbf{33.85}}  &  {0.8798}          &  {0.9218}          &  {0.5586}              &  {4.9961} \\
 {0.005}          &  {0.9445}            &  {$2.03\times10^{-5}$}      &  {33.98}           &  {0.8858}          &  {0.9264}          &  {0.5743}              &  {5.1987}
                            &  {0.9923}            &  {$3.03\times10^{-6}$}      &  {33.82}           &  {0.8785}          &  {0.9216}          &  {0.5579}              &  {\textbf{5.0183}} \\
 {0.01}           &  {0.9941}            &  {$2.78\times10^{-6}$}      &  {34.08}           &  {0.8863}          &  {0.9263}          &  {0.5746}              &  {5.2217}
                            &  {0.9925}            &  {$2.87\times10^{-6}$}      &  {33.84}           &  {0.8797}          &  {0.9218}          &  {0.5557}              &  {5.0175} \\
 {0.05}           &  {\textbf{0.9978}}   &  {$1.04\times10^{-6}$}      &  {\textbf{34.13}}  &  {\textbf{0.8881}} &  {\textbf{0.9267}} &  {\textbf{0.5761}}     &  {5.2205}
                            &  {0.9979}            &  {$1.26\times10^{-6}$}      &  {33.81}           &  {0.8800}          &  {0.9218}          &  {\textbf{0.5590}}     &  {5.0143} \\
 {0.1}            &  {\textbf{0.9978}}   &  {$\bf{8.39\times10^{-7}}$} &  {34.04}           &  {0.8864}          &  {0.9263}          &  {0.5737}              &  {\textbf{5.2588}}
                            &  {\textbf{0.9987}}   &  {$\bf{7.79\times10^{-7}}$} &  {33.70}           &  {\textbf{0.8801}} &  {0.9223}          &  {0.5572}              &  {4.9969} \\
 {0.5}            &  {0.9965}            &  {$1.12\times10^{-6}$}      &  {33.91}           &  {0.8835}          &  {0.9258}          &  {0.5686}              &  {5.2132}
                            &  {0.9963}            &  {$1.29\times10^{-6}$}      &  {33.50}           &  {0.8745}          &  {0.9207}          &  {0.5500}              &  {4.9646} \\
\bottomrule[1.5pt]
    \end{tabular*}
\end{table*}

\begin{figure*}[htbp]
\centering
\includegraphics[width=7in]{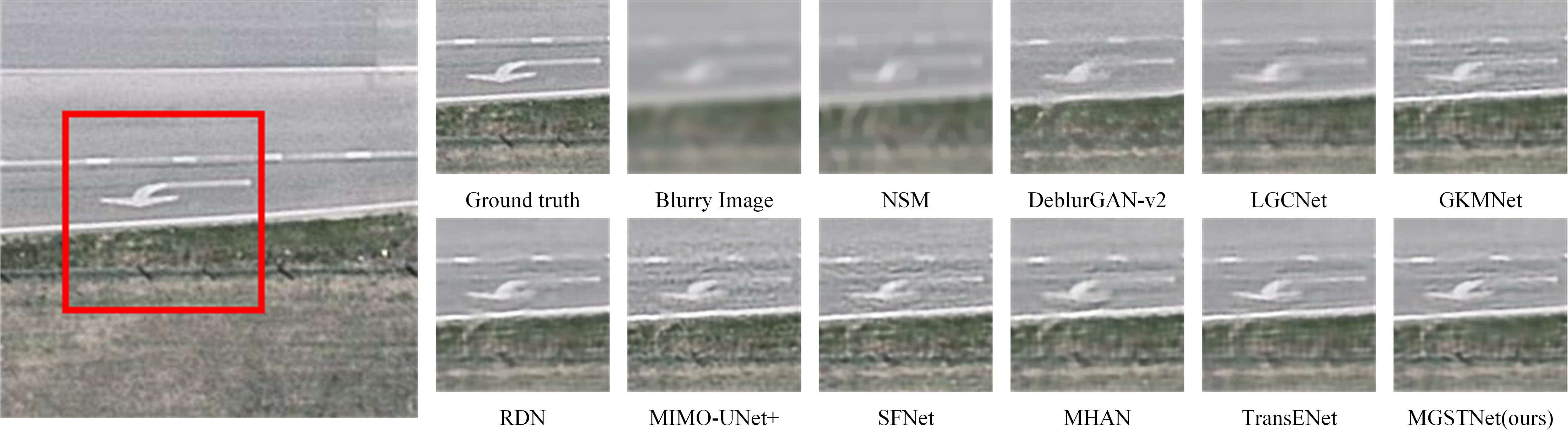}
\caption{Visual comparison of different deblurring methods on AIRS datasets.}
\label{fig4}
\end{figure*}

\begin{figure*}[htbp]
\centering
\includegraphics[width=7in]{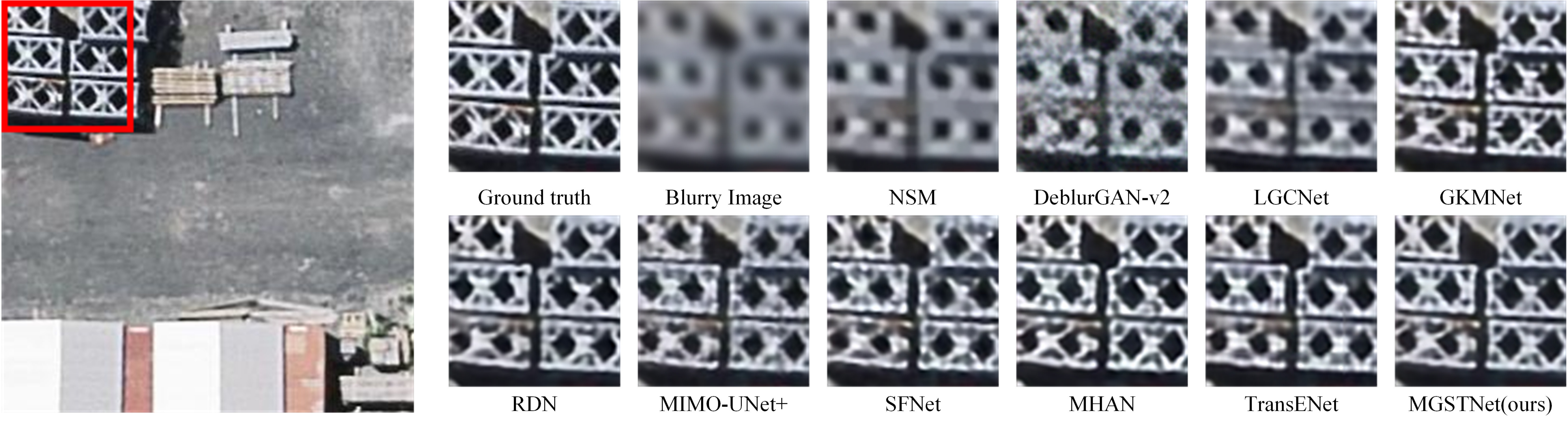}
\caption{Visual comparison of different deblurring methods on AIRS datasets with small perturbation noise.}
\label{fig5}
\end{figure*}

\subsection{Ablation Study and Architecture {Evalution}}
In this part, we conducted a series of experiments on the AIRS dataset to determine the optimal configuration for the proposed MGSTNet in image deblurring. These experiments focused on different components such as the number of scales in $F(\cdot)$ and $\tilde{F}(\cdot)$, the number of RCABs in $G(\cdot)$, $\tilde{G}(\cdot)$, encoder, and decoder, as well as the number of stages. For a easier evaluation of the reconstruction performance, we performed all ablation studies with a stage number of $k = 1$, except the ablation study on the number of stages.

\subsubsection{\textbf{Number of RCABs} in $\bf{G_{k}(\cdot)}$ \textbf{and} $\bf{\tilde{G}_{k}(\cdot)}$}
The impact of depth on the performance of blur kernel deep prior networks is demonstrated in Fig. \ref{fig2}(a) and Table \ref{table1}. The results clearly show that increasing the number of RCABs from 1 to 2 leads to an improvement in image recovery. However, additional increaments in depth do not result in further gains. This indicates that a simpler network structure can effectively learn the blur kernel deep prior.

\subsubsection{\textbf{The impact of KPMM}}
To demonstrate the impact of the KPMM module on the performance of blur kernel reconstruction, we assess the kernels using three metrics: mean squared error (MSE), root mean squared error (RMSE), and maximum normalized convolution (MNC) \cite{hu2012good} in Table \ref{table1b}. After the blur kernel undergoes the gradient descent step, an initial estimate $\bm{s}$ of the blur kernel is obtained, but it is not accurate. However, after passing through the KPMM module, the error of the blur kernel is significantly reduced. The results in Table \ref{table1b} indicate that the KPMM module can effectively learn the depth prior of the blur kernel.

\subsubsection{\textbf{Number of scales in} $\bf{F_{k}(\cdot)}$ \textbf{and} $\bf{\tilde{F}_{k}(\cdot)}$}
In this part, we investigate the impact of multi-scale and single-scale thresholds in our network, as illustrated in Fig. \ref{fig2}(b) and summarized in Table \ref{table2}. Our findings reveal that the variant with multi-scale shrinkage thresholds consistently outperforms the one with single-scale thresholds across all metrics. Specifically, excluding the multi-scale shrinkage thresholds results in a decrease in PSNR by 0.35 dB. These results clearly demonstrate that the proposed method, which incorporates a multi-scale prior, consistently outperforms the single-scale approach in terms of deblur performance. This emphasizes the importance of integrating multi-scale thresholding techniques into network design to enhance the quality of image reconstruction.

\subsubsection{\textbf{Number of RCABs in Encoder and Decoder}} In this section, we investigate the advantages of integrating multiple RCABs in both the Encoder and Decoder parts of the MGSTNet framework. As shown in Fig. \ref{fig2}(c) and {in} Table \ref{table3}, the performance of the framework improves as the number of RCABs increases. This result confirms the effectiveness of the image prior network design. To achieve a more accurate reconstruction of the geometric prior of the image, our MGSTNet network employs two RCABs.

\subsubsection{\textbf{Number of stages}} We further examine the relationship between the number of stages and the reconstruction performance by varying the stage number to 1, 2, and 3. The results, shown in Fig. \ref{fig2}(d) and Table \ref{table4}, indicate that the performance improves as the number of stages increases, thus validating the effectiveness of the iterative network design. Based on these findings, we select three stages for the MGSTNet framework to achieve the best reconstruction performance.

\subsubsection{\textbf{Sensitivity analysis of the loss weight $\alpha$}}
Experiments were carried out on the AIRS dataset to investigate the appropriate loss weight $\alpha$. Various values of $\alpha$ were tested with MGSTNet. The results in Table \ref{table7} indicate that the optimal performance for blur kernel reconstruction is achieved when $\alpha$ is set to 0.05 and 0.1. Similarly, the best performance for image deblurring reconstruction is observed at approximately $\alpha = 0.05$. Therefore, we set $\alpha$ to 0.05 throughout all experiments.

\subsection{Experiments on AIRS Dataset}
To demonstrate the superiority of the proposed method, we compare it with several state-of-the-art methods on the synthetic AIRS dataset.

\begin{figure*}[htbp]
\centering
\includegraphics[width=7in]{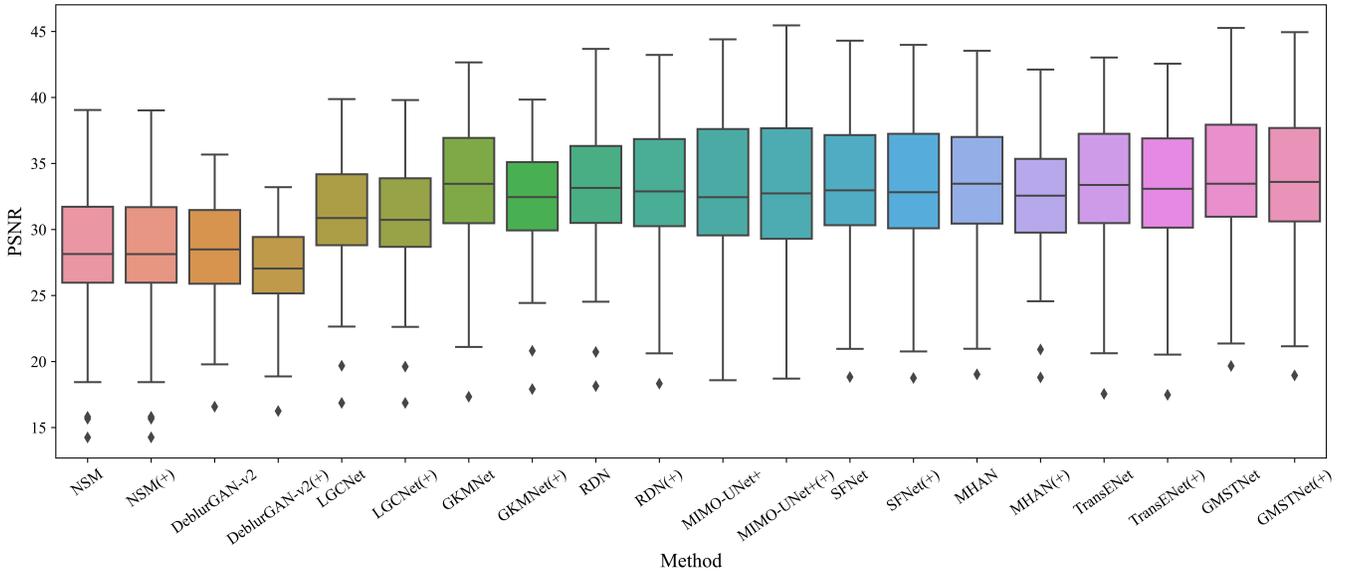}
\caption{{The PSNR box-plot of different deblurring methods on AIRS. '(+)' represents the deblurring performance experiment with noise.}}
\label{fig3}
\end{figure*}

\begin{table*}
    \caption{Performance comparisons of various deblurring methods on the AIRS datasets in terms of the interpretability, the estimation of the degradation kernel, the number of parameters, {the average PSNR, SSIM, FSIM, VIF, and IFC}. The semi-kernel estimation strategy in GKMNet denoted by \CBmark utilizes a weighted combination of pre-defined Gaussian kernels for its kernel estimation, while NSM is a non-learning deblurring technique.}
    \label{table5}
    \renewcommand\arraystretch{1.1}
    \setlength\tabcolsep{2pt}
    \centering
    \begin{tabular*}{\hsize}{@{}@{\extracolsep{\fill}}lccccccccccccc@{}}
    \toprule[1.5pt]
\multirow{2}{*}{Method} & \multirow{2}{*}{Interpretability} &\multirow{2}{*}{Kernel estimated}& \multirow{2}{*}{Parameters}
 & \multicolumn{5}{c}{AIRS} & \multicolumn{5}{c}{AIRS(+)} \\
 \cmidrule(r){5-9} \cmidrule(r){10-14}
& & & & PSNR&SSIM&{FSIM}&{VIF}&{IFC}   & PSNR&SSIM&{FSIM}&{VIF}&{IFC} \\
\midrule[0.95pt]
NSM \cite{krishnan2011blind}            & \cmark & \cmark  & \textbf{--}     & 28.24 & 0.6903 & {0.7439} & {0.2550} & {2.0360} & 28.23 & 0.6897 & {0.7440} & {0.2539} & {2.0280} \\
DeblurGAN-v2 \cite{kupyn2019deblurgan}  & \xmark & \xmark  & $60.90\text{M}$ & 28.53 & 0.7835 & {0.8691} & {0.4257} & {3.1947} & 27.17 & 0.6686 & {0.8468} & {0.2918} & {2.0542} \\
{LGCNet} \cite{7937881}                   & {\xmark} & {\xmark}  & {$0.19\text{M}$}  & {30.98} & {0.8211} & {0.8798} & {0.4384} & {3.2458}                                                 & {30.88} & {0.8170} & {0.8775} & {0.4302} & {3.1796} \\
GKMNet \cite{quan2021gaussian}          & \cmark & \CBmark & $1.41\text{M}$  & 33.21 & 0.8667 & {0.9177} & {0.5588} & {4.7611} & 32.37 & 0.8681 & {0.9209} &{\textbf{0.5612}} & {4.4824} \\
RDN  \cite{zhang2020residual}           & \xmark & \xmark  & $21.98\text{M}$ & 33.21 & 0.8782 & {0.9196} & {0.5644} & {4.8520} & 33.10 & 0.8708 & {0.9167} & {0.5434} & {4.6961} \\
MIMO-UNet+ \cite{cho2021rethinking}     & \xmark & \xmark  & $16.11\text{M}$ & 33.19 & 0.8644 & {0.9199} & {0.5506} & {4.8974} & 33.13 & 0.8538 & {0.9137} & {0.5308} & {4.7728} \\
SFNet \cite{cui2022selective}           & \xmark & \xmark  & $13.27\text{M}$ & 33.37 & 0.8769 & {0.9247} & {0.5698} & {5.0076} & 33.29 & 0.8748 & {\textbf{0.9241}} & {0.5601} & {5.0094} \\
{MHAN}  \cite{9151234}                    & {\xmark} & {\xmark}  & {$11.42\text{M}$} & {33.42} & {0.8735} & {0.9172} & {0.5465} & {4.8347}                                                 & {32.37} & {0.8584} & {0.9069} & {0.4908} & {4.2222} \\
{TransENet}  \cite{9654169}               & {\xmark} & {\xmark}  & {$37.16\text{M}$} & {33.32} & {0.8631} & {0.9120} & {0.5317} & {4.4925}                                                 & {33.03} & {0.8546} & {0.9048} & {0.5154} & {4.3655} \\
MGSTNet (Ours)                          & \cmark & \cmark  & $15.23\text{M}$ & \textbf{34.13}&\textbf{0.8881}&{\textbf{0.9267}}&{\textbf{0.5761}} & {\textbf{5.2205}} & \textbf{33.81} &\textbf{0.8800} & {0.9218} & {0.5590} & {\textbf{5.0143}} \\
\bottomrule[1.5pt]
    \end{tabular*}
\end{table*}

\begin{figure*}[htbp]
\centering
\includegraphics[width=7in]{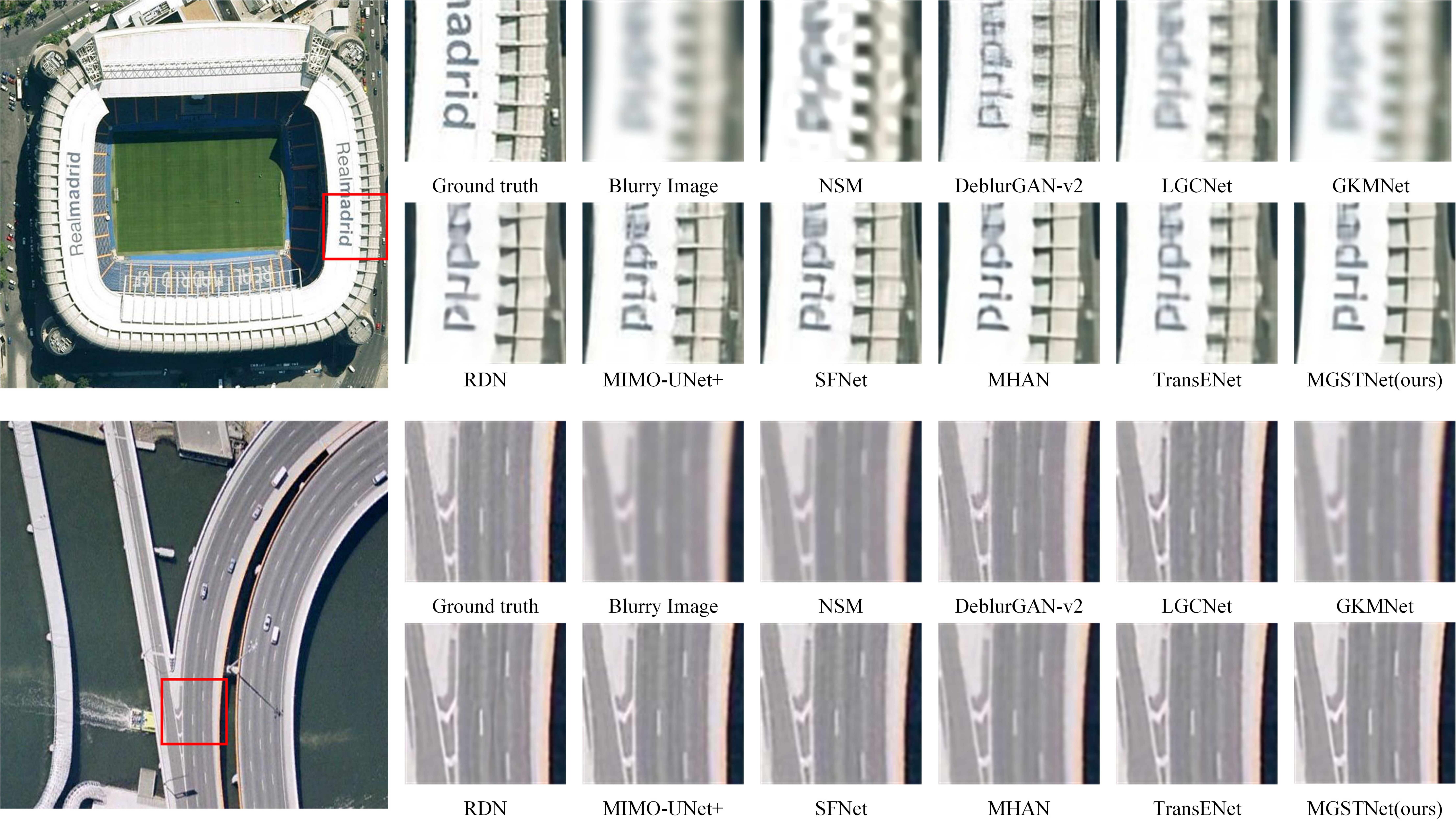}
\caption{{Visual comparison for different deblurring methods on WHU-RS19 dataset.}}
\label{fig6}
\end{figure*}

\begin{table*}
    \caption{Comparison of different deblurring models on various scene classes of WHU-RS19 (PSNR ($dB$)/SSIM/FSIM/VIF/IFC). The best and second-best results are highlighted in Bold font and underlined ones, respectively.}
    \label{table6}
    \renewcommand\arraystretch{1.1}
    \setlength\tabcolsep{2pt}
    \centering
    \begin{tabular*}{\hsize}{@{}@{\extracolsep{\fill}}lcccc@{}}
    \toprule[1.5pt]
\multirow{1}{*}{Method} & \multicolumn{4}{c}{Different scene classes of WHU-RS19}  \\
    \midrule[1.0pt]
\multirow{2}{*}{ } & \multicolumn{1}{c}{Airport} & \multicolumn{1}{c}{Beach} & \multicolumn{1}{c}{Bridge} & \multicolumn{1}{c}{Commercial} \\
\midrule[0.95pt]
& PSNR/SSIM/{FSIM}/{VIF}/{IFC} & PSNR/SSIM/{FSIM}/{VIF}/{IFC} & PSNR/SSIM/{FSIM}/{VIF}/{IFC}   & PSNR/SSIM/{FSIM}/{VIF}/{IFC} \\
 \cmidrule(r){2-5}
 {NSM} \cite{krishnan2011blind}            &  {25.32}/{0.7574}/{0.9011}/{0.336}/{2.930} &  {40.64}/{0.9547}/{0.9754}/{0.615}/{1.923} &  {25.51}/{0.7598}/{0.8259}/{0.241}/{1.419} &  {22.12}/{0.6779}/{0.8773}/{0.270}/{2.883} \\
DeblurGAN-v2 \cite{kupyn2019deblurgan}  & 29.70/0.8364/{0.9471}/{0.496}/{4.869} & 38.65/0.9499/{0.9737}/{0.662}/{2.023} & 32.27/0.8921/{0.9471}/{0.470}/{2.981} & 27.48/0.8196/{0.9509}/{0.488}/{5.762} \\
 {LGCNet} \cite{7937881}      &  {29.29}/{0.8377}/{0.9450}/{0.485}/{4.427} &  {40.08}/{0.9558}/{0.9740}/{0.695}/{2.119} &  {33.73}/{0.9030}/{0.9523}/{0.486}/{2.962} &  {27.47}/{0.8218}/{0.9491}/{0.470}/{5.430} \\
GKMNet \cite{quan2021gaussian}          & 27.36/0.7681/{0.8999}/{0.408}/{3.969} & 36.75/0.9477/{0.9613}/{0.632}/{2.086} & 31.53/0.8709/{0.9132}/{0.400}/{2.743} & 25.43/0.7225/{0.8946}/{0.379}/{4.562} \\
RDN  \cite{zhang2020residual}           & 31.34/\underline{0.8719}/{0.9638}/{\underline{0.575}}/{\underline{6.093}} & 40.96/0.9595/{0.9801}/{0.733}/{2.380} & 35.33/\underline{0.9181}/{0.9665}/{\textbf{0.549}}/{3.741} & \underline{29.27}/\underline{0.8586}/{0.9653}/{\underline{0.565}}/{\underline{7.364}} \\
MIMO-UNet+ \cite{cho2021rethinking}     & 30.13/0.8603/{0.9652}/{0.547}/{5.155} & 41.95/0.9641/{0.9870}/{0.719}/{2.435} & 34.26/0.9082/{0.9667}/{0.512}/{3.003} & 27.86/0.8433/{0.9649}/{0.536}/{6.347} \\
SFNet \cite{cui2022selective}           & 30.61/0.8706/{\underline{0.9689}}/{0.557}/{5.258} & 41.16/0.9634/{0.9856}/{\underline{0.737}}/{2.407} & 34.74/0.9115/{\underline{0.9689}}/{0.521}/{3.045} & 28.32/0.8559/{\textbf{0.9687}}/{0.547}/{6.481} \\
 {MHAN}  \cite{9151234}       &  {31.31}/{0.8703}/{0.9626}/{0.547}/{5.876} &  {\textbf{43.88}}/{0.9707}/{\underline{0.9878}}/{0.688}/{\underline{2.829}} &  {35.53}/{0.9145}/{0.9640}/{0.504}/{\underline{3.845}} &
 {29.10}/{0.8553}/{0.9636}/{0.540}/{6.935} \\
 {TransENet}  \cite{9654169}  &  {\underline{31.54}}/{0.8711}/{0.9640}/{0.564}/{5.873} &  {42.38}/{\underline{0.9711}}/{0.9771}/{0.711}/{2.524} &  {\underline{35.72}}/{0.9168}/{0.9663}/{0.526}/{3.412} &  {\underline{29.27}}/{0.8540}/{0.9644}/{0.553}/{7.015} \\
MGSTNet (Ours)                          & \textbf{32.10}/\textbf{0.8824}/{\textbf{0.9698}}/{\textbf{0.584}}/{\textbf{6.481}} & \underline{43.69}/\textbf{0.9733}/{\textbf{0.9901}}/{\textbf{0.740}}/{\textbf{2.982}} & \textbf{36.13}/\textbf{0.9222}/{\textbf{0.9699}}/{\underline{0.540}}/{\textbf{4.232}} & \textbf{29.68}/\textbf{0.8655}/{\underline{0.9685}}/{\textbf{0.569}}/{\textbf{7.493}} \\
\midrule[0.95pt]
\multirow{2}{*}{ } & \multicolumn{1}{c}{Desert} & \multicolumn{1}{c}{Farmland} & \multicolumn{1}{c}{FootballField} & \multicolumn{1}{c}{Forest} \\
\midrule[0.95pt]
 {NSM} \cite{krishnan2011blind}            &  {37.63}/{0.9020}/{0.9500}/{0.431}/{2.070} &  {31.71}/{0.7832}/{0.8764}/{0.392}/{2.187} &  {25.08}/{0.7651}/{0.8991}/{0.368}/{3.067} &  {26.97}/{0.6615}/{0.8838}/{0.277}/{2.659} \\
DeblurGAN-v2 \cite{kupyn2019deblurgan}  & 38.31/0.9166/{0.9683}/{0.506}/{2.551} & 35.98/0.8726/{0.9498}/{0.561}/{3.755} & 31.61/0.8774/{0.9629}/{0.583}/{5.510} & 28.67/0.7451/{0.9349}/{0.421}/{4.363} \\
 {LGCNet} \cite{7937881}      &
 {39.27}/{0.9223}/{0.9689}/{\underline{0.529}}/{2.663} &  {35.24}/{0.8670}/{0.9452}/{0.542}/{3.379} &  {30.70}/{0.8704}/{0.9579}/{0.556}/{4.871} &  {29.67}/{0.7634}/{0.9412}/{0.425}/{4.462} \\
GKMNet \cite{quan2021gaussian}          & 35.18/0.9013/{0.9401}/{0.473}/{2.384} & 34.86/0.8520/{0.9186}/{0.504}/{3.683} & 29.36/0.8217/{0.9221}/{0.501}/{4.723} & 27.90/0.6452/{0.8627}/{0.322}/{3.568} \\
RDN  \cite{zhang2020residual}           & 40.14/\underline{0.9289}/{0.9761}/{\textbf{0.566}}/{\underline{3.333}} & 37.67/0.8903/{0.9618}/{\underline{0.602}}/{4.692} & 33.70/0.9033/{0.9742}/{\underline{0.650}}/{\underline{7.008}} & 30.62/0.7967/{0.9556}/{\underline{0.489}}/{\underline{5.595}} \\
MIMO-UNet+ \cite{cho2021rethinking}     & 38.94/0.9188/{0.9759}/{0.484}/{2.371} & 36.41/0.8804/{0.9628}/{0.573}/{3.740} & 31.99/0.8918/{0.9746}/{0.619}/{5.633} & 29.64/0.7909/{0.9599}/{0.462}/{4.835} \\
SFNet \cite{cui2022selective}           & 39.03/0.9191/{0.9767}/{0.492}/{2.386} & 36.83/0.8872/{\underline{0.9667}}/{0.585}/{3.829} & 32.34/0.8977/{\underline{0.9769}}/{0.627}/{5.687} & 29.89/0.7989/{\underline{0.9617}}/{0.467}/{4.883} \\
 {MHAN}  \cite{9151234}       &  {\underline{40.26}}/{0.9255}/{0.9743}/{0.481}/{3.174} &  {\underline{38.35}}/{0.8871}/{0.9588}/{0.557}/{\underline{4.914}} &  {\underline{34.31}}/{\underline{0.9038}}/{0.9739}/{0.629}/{6.951} &  {\underline{30.73}}/{\underline{0.8067}}/{0.9609}/{0.488}/{5.566} \\
 {TransENet}  \cite{9654169}  &  {40.04}/{0.9285}/{\underline{0.9776}}/{0.517}/{2.993} &  {38.33}/{\underline{0.8920}}/{0.9625}/{0.581}/{4.594} &  {34.27}/{0.9020/{0.9741}}/{0.641}/{6.662} &  {30.69}/{0.8041/{0.9591}}/{0.487}/{5.402} \\
MGSTNet (Ours)                          & \textbf{40.71}/\textbf{0.9309}/{\textbf{0.9784}}/{0.516}/{\textbf{3.479}} & \textbf{38.46}/\textbf{0.8987}/{\textbf{0.9681}}/{\textbf{0.607}}/{\textbf{4.938}} & \textbf{34.64}/\textbf{0.9119}/{\textbf{0.9778}}/{\textbf{0.663}}/{\textbf{7.440}} & \textbf{30.93}/\textbf{0.8109}/{\textbf{0.9624}}/{\textbf{0.499}}/{\textbf{5.783}} \\
\midrule[0.95pt]
\multirow{2}{*}{ } & \multicolumn{1}{c}{Industrial} & \multicolumn{1}{c}{Meadow} & \multicolumn{1}{c}{Mountain} & \multicolumn{1}{c}{Park} \\
\midrule[0.95pt]
 {NSM} \cite{krishnan2011blind}            &  {23.86}/{0.7042}/{0.8774}/{0.278}/{2.682} &  {35.04}/{0.8370}/{0.9246}/{0.408}/{2.568} &  {23.72}/{0.6304}/{0.8863}/{0.267}/{2.880} &  {26.57}/{0.7282}/{0.9047}/{0.315}/{2.823} \\
DeblurGAN-v2 \cite{kupyn2019deblurgan}  & 28.26/0.8183/{0.9459}/{0.463}/{4.775} & 36.11/0.8684/{0.9538}/{0.529}/{3.496} & 25.08/0.6298/{0.9332}/{0.398}/{4.667} & 28.94/0.7881/{0.9398}/{0.446}/{4.363} \\
 {LGCNet} \cite{7937881}      &  {27.57}/{0.8123}/{0.9390}/{0.426}/{4.183} &  {36.74}/{0.8780}/{0.9574}/{0.552}/{3.391} &  {26.73}/{0.7428}/{0.9412}/{0.417}/{4.935} &  {28.74}/{0.7881}/{0.9358}/{0.425}/{4.009} \\
GKMNet \cite{quan2021gaussian}          & 25.80/0.7199/{0.8822}/{0.350}/{3.606} & 35.62/0.8536/{0.9262}/{0.455}/{3.348} & 25.08/0.6298/{0.8749}/{0.321}/{3.920} & 27.16/0.7123/{0.8849}/{0.346}/{3.447} \\
RDN  \cite{zhang2020residual}           & 29.51/0.8545/{0.9607}/{\underline{0.529}}/{\underline{5.768}} & 38.46/\textbf{0.8949}/{\underline{0.9713}}/{\textbf{0.586}}/{\textbf{4.444}} & \underline{27.64}/0.7764/{0.9556}/{\underline{0.489}}/{\textbf{6.452}} & 30.16/\underline{0.8263}/{0.9558}/{\textbf{0.511}}/{\underline{5.339}} \\
MIMO-UNet+ \cite{cho2021rethinking}     &
28.84/0.8441/{0.9620}/{0.512}/{5.325} & 37.33/0.8848/{\textbf{0.9719}}/{0.550}/{3.542} & 26.49/0.7639/{0.9567}/{0.461}/{5.532} & 29.37/0.8124/{0.9562}/{0.485}/{4.673}\\
SFNet \cite{cui2022selective}           & 29.42/0.8573/{\underline{0.9663}}/{0.527}/{5.483} & 37.39/0.8831/{0.9705}/{0.538}/{3.517} & 26.81/0.7747/{\textbf{0.9602}}/{0.469}/{5.614} & 29.81/0.8230/{\textbf{0.9601}}/{0.494}/{4.782} \\
 {MHAN}  \cite{9151234}       &  {29.75}/{\underline{0.8581}}/{0.9618}/{0.514}/{5.616} &  {\underline{38.56}}/{0.8845}/{0.9653}/{0.506}/{4.293} &  {27.62}/{0.7767}/{0.9566}/{0.479}/{6.230} &  {30.00}/{0.8187}/{0.9531}/{0.473}/{4.939} \\
 {TransENet}  \cite{9654169}  &  {\underline{29.92}}/{0.8512}/{0.9600}/{0.522}/{5.727} &  {\textbf{38.62}}/{\underline{0.8928}}/{0.9692}/{0.548}/{4.212} &  {27.62}/{\underline{0.7770}}/{0.9564}/{0.485}/{6.151} &  {\underline{30.35}}/{0.8212}/{0.9549}/{0.498}/{5.165} \\
MGSTNet (Ours)                          & \textbf{30.41}/\textbf{0.8675}/{\textbf{0.9667}}/{\textbf{0.548}}/{\textbf{6.145}} & 38.55/0.8890/{0.9671}/{\underline{0.562}}/{\underline{4.440}} & \textbf{27.74}/\textbf{0.7856}/{\underline{0.9601}}/{\textbf{0.494}}/{\underline{6.437}} & \textbf{30.71}/\textbf{0.8310}/{\underline{0.9597}}/{\underline{0.510}}/{\textbf{5.496}} \\
\midrule[0.95pt]
\multirow{2}{*}{ } & \multicolumn{1}{c}{Parking} & \multicolumn{1}{c}{Pond} & \multicolumn{1}{c}{Port} & \multicolumn{1}{c}{RailwayStation} \\
\midrule[0.95pt]
 {NSM} \cite{krishnan2011blind}            &  {23.54}/{0.7487}/{0.8734}/{0.270}/{2.454} &  {29.59}/{0.8392}/{0.9145}/{0.341}/{2.265} &  {24.74}/{0.7969}/{0.8981}/{0.297}/{2.505} &  {23.41}/{0.6578}/{0.8725}/{0.275}/{2.733} \\
DeblurGAN-v2 \cite{kupyn2019deblurgan}  & 28.22/0.8551/{0.9452}/{0.462}/{4.416} & 31.77/0.8663/{0.9424}/{0.460}/{3.127} & 28.12/0.8586/{0.9481}/{0.465}/{4.009} & 27.41/0.7707/{0.9402}/{0.445}/{4.927} \\
 {LGCNet} \cite{7937881}      &  {27.31}/{0.8448}/{0.9344}/{0.421}/{3.816} &  {32.18}/{0.8733}/{0.9428}/{0.464}/{2.967} &  {28.82}/{0.8671}/{0.9469}/{0.457}/{3.967} &  {27.05}/{0.7648}/{0.9341}/{0.421}/{4.522} \\
GKMNet \cite{quan2021gaussian}          & 25.32/0.7585/{0.8702}/{0.338}/{3.320} & 30.88/0.8308/{0.8998}/{0.392}/{2.848} & 26.66/0.8010/{0.8936}/{0.365}/{3.382} & 25.53/0.6755/{0.8783}/{0.356}/{4.017} \\
RDN  \cite{zhang2020residual}           & 29.37/0.8837/{0.9585}/{0.521}/{5.286} & 34.00/\underline{0.8940}/{0.9596}/{\underline{0.523}}/{4.038} & 30.60/0.8927/{0.9650}/{\underline{0.543}}/{\underline{5.197}} & 28.67/\underline{0.8086}/{0.9565}/{\textbf{0.513}}/{\textbf{6.135}} \\
MIMO-UNet+ \cite{cho2021rethinking}     & 28.84/0.8757/{0.9613}/{0.510}/{4.860} & 32.79/0.8852/{0.9610}/{0.492}/{3.197} & 29.33/0.8765/{0.9616}/{0.514}/{4.374} & 27.60/0.7979/{0.9570}/{0.488}/{5.297} \\
SFNet \cite{cui2022selective}           & 29.76/0.8894/{\textbf{0.9667}}/{0.532}/{5.086} & 33.20/0.8891/{\underline{0.9632}}/{0.497}/{3.241} & 29.95/0.8861/{\textbf{0.9664}}/{0.529}/{4.496} & 28.04/0.8066/{\textbf{0.9607}}/{0.495}/{5.396} \\
 {MHAN}  \cite{9151234}       &  {\textbf{30.79}}/{\underline{0.8969}}/{0.9657}/{\underline{0.540}}/{\textbf{5.667}} &  {\underline{34.30}}/{0.8936}/{0.9595}/{0.496}/{\underline{4.105}} &  {30.85}/{\underline{0.8944}}/{0.9653}/{0.527}/{5.104} &  {28.73}/{0.8003}/{0.9528}/{0.492}/{5.946} \\
 {TransENet}  \cite{9654169}  &  {30.43}/{0.8881}/{0.9619}/{0.528}/{5.345} &  {34.19}/{0.8924}/{0.9593}/{0.508}/{3.663} &  {\underline{30.88}}/{0.8921}/{0.9656}/{0.533}/{4.852} &  {\underline{28.88}}/{0.8044}/{0.9557}/{0.503}/{5.939} \\
MGSTNet (Ours)                          & \underline{30.51}/\textbf{0.8980}/{\underline{0.9662}}/{\textbf{0.549}}/{\underline{5.660}} & \textbf{34.59}/\textbf{0.8994}/{\textbf{0.9642}}/{\textbf{0.525}}/{\textbf{4.553}} & \textbf{30.95}/\textbf{0.8966}/{\underline{0.9659}}/{\textbf{0.548}}/{\textbf{5.310}} & \textbf{28.89}/\textbf{0.8105}/{\underline{0.9574}}/{\underline{0.512}}/{\underline{6.081}} \\
\midrule[0.95pt]
\multirow{2}{*}{ } & \multicolumn{1}{c}{Residential} & \multicolumn{1}{c}{River} & \multicolumn{1}{c}{Viaduct} & \multicolumn{1}{c}{Average} \\
\midrule[0.95pt]
 {NSM} \cite{krishnan2011blind}            &  {22.49}/{0.7117}/{0.8824}/{0.261}/{2.805} &  {27.96}/{0.7784}/{0.9270}/{0.391}/{3.478} &  {23.43}/{0.7149}/{0.8877}/{0.313}/{3.218} &  {27.32}/{0.7583}/{0.8968}/{0.334}/{2.618} \\
DeblurGAN-v2 \cite{kupyn2019deblurgan}  & 27.26/0.8415/{0.9552}/{0.479}/{5.464} & 31.04/0.8320/{0.9559}/{0.556}/{5.523} & 28.55/0.8389/{0.9549}/{0.518}/{5.848} & 30.75/0.8404/{0.9500}/{0.495}/{4.357} \\
 {LGCNet} \cite{7937881}      &  {27.12}/{0.8410}/{0.9521}/{0.459}/{5.112} &  {31.26}/{0.8407}/{0.9549}/{0.542}/{5.211} &  {28.14}/{0.8384}/{0.9522}/{0.495}/{5.354} &  {30.89}/{0.8439}/{0.9487}/{0.488}/{4.111} \\
GKMNet \cite{quan2021gaussian}          & 24.38/0.7425/{0.8986}/{0.345}/{3.934} & 30.39/0.7852/{0.9173}/{0.476}/{4.856} & 26.32/0.7491/{0.9033}/{0.417}/{4.748} & 29.03/0.7783/{0.9023}/{0.409}/{3.654} \\
RDN  \cite{zhang2020residual}           & 28.85/0.8797/{0.9699}/{0.567}/{6.866} & 32.87/\underline{0.8648}/{0.9683}/{\underline{0.614}}/{\underline{6.790}} & 30.24/\underline{0.8761}/{0.9707}/{\underline{0.600}}/{\underline{7.309}} & 32.60/\underline{0.8727}/{0.9651}/{\underline{0.565}}/{\underline{5.488}} \\
MIMO-UNet+ \cite{cho2021rethinking}     & 28.13/0.8674/{0.9704}/{0.552}/{6.453} & 31.42/0.8520/{0.9694}/{0.574}/{5.574} & 29.14/0.8653/{0.9719}/{0.575}/{6.451} & 31.59/0.8623/{0.9662}/{0.535}/{4.692} \\
SFNet \cite{cui2022selective}           & 28.78/0.8797/{\underline{0.9741}}/{0.566}/{6.635} & 31.68/0.8575/{\underline{0.9711}}/{0.578}/{5.620} & 29.44/0.8723/{\underline{0.9734}}/{0.579}/{6.489} & 31.95/0.8697/{\underline{0.9688}}/{0.544}/{4.773} \\
 {MHAN}  \cite{9151234}       &  {\underline{29.28}}/{\underline{0.8865}}/{0.9724}/{\underline{0.569}}/{\underline{6.923}} &  {32.92}/{0.8632}/{0.9686}/{0.581}/{6.483} &  {30.17}/{0.8727}/{0.9689}/{0.578}/{7.040} &  {\underline{32.95}}/{0.8726}/{0.9651}/{0.536}/{5.409} \\
 {TransENet}  \cite{9654169}  &  {29.12}/{0.8804}/{0.9705}/{0.566}/{6.800} &  {\underline{33.14}}/{0.8642}/{0.9686}/{0.603}/{6.543} &  {\underline{30.45}}/{0.8744}/{0.9705}/{0.594}/{7.136} &  {32.93}/{0.8726}/{0.9652}/{0.551}/{5.285} \\
MGSTNet (Ours)                          & \textbf{29.50}/\textbf{0.8898}/{\textbf{0.9742}}/{\textbf{0.583}}/{\textbf{7.127}} & \textbf{33.52}/\textbf{0.8714}/{\textbf{0.9720}}/{\textbf{0.617}}/{\textbf{7.113}} & \textbf{31.02}/\textbf{0.8855}/{\textbf{0.9748}}/{\textbf{0.613}}/{\textbf{7.657}} & \textbf{33.29}/\textbf{0.8800}/{\textbf{0.9691}}/{\textbf{0.567}}/{\textbf{5.751}} \\
\bottomrule[1.5pt]
    \end{tabular*}
\end{table*}

Table \ref{table5} first presents that several models, such as DeblurGAN-v2, LGCNet, RDN, MIMO-UNet+, SFNet, MHAN, and TransENet, belong to the category of black box models, indicating their lack of transparency and interpretability. In contrast, NSM and MGSTNet are built on model-driven theory to develop blur kernel estimation learning modules. GKMNet, on the other hand, utilizes a weighted combination of predefined Gaussian kernels for kernel estimation. It should be noted that, except for NSM, all nine techniques rely on deep learning, which is known for its superior performance compared to traditional image deblurring methods. MGSTNet has a larger number of parameters compared to GKMNet due to the inclusion of image/kernel prior features necessary for image deblurring. However, MGSTNet still maintains a reasonable parameter count and achieves the highest performance in image blind deblurring among all compared approaches.

Table \ref{table5} also illustrates a quantitative comparison of the deblurring effects in a specific experimental setup, where the performance with small perturbation noise is denoted AIRS(+). The introduction of noise deteriorates the quality of the original image, making it difficult to extract feature information. Therefore, it is crucial for image deblurring methods to have a strong capability to reconstruct details. Our method outperforms other state-of-the-art approaches in both remote sensing image deblurring tasks, showing an average improvement of 0.71 dB on the AIRS dataset. The performance of the different methods decreases as a result of the addition of noise, with the NSM remaining relatively stable. However, our method still outperforms other deblurring methods in terms of performance.

Fig. \ref{fig4} and Fig. \ref{fig5} display the deblurring reconstructions obtained using different methods. Our approach successfully eliminates the degradation caused by blur, resulting in a sharper image that closely resembles the ground truth in terms of visual perception compared to other methods. Conversely, the deblurring results of NSM, DeblurGAN-v2, LGCNet, and GKMNet exhibit noticeable blurring, while the deblurring results of DeblurGAN-v2, GKMNet, MIMO-UNet+, and SFNet contain visible artifacts. The comparison shown in Figs. \ref{fig4} and \ref{fig5} demonstrates that our MGSTNet effectively recovers detailed information from the deblurred image, making its performance superior to the other methods considered. Furthermore, MGSTNet also performs well in handling deblurring problems involving noise.

The box-plot can be used to evaluate the stability of the deblurring effect. It consists of the minimum, first quartile (Q1), median, third quartile (Q3), and maximum values of the data. The Interquartile Range (IQR) defined by $IQR = Q3 - Q1$ is used to measure the discrete degree of the data. Outliers are defined as data points that are more than 1.5 IQR away from either the Q1 or Q3. Fig. \ref{fig3} shows that all deblurring methods, except for MIMO-UNet+, exhibit outliers, which are represented by diamonds on the lower side of Fig. \ref{fig3}. The reason for this is that MIMO-UNet+ has an excessively high interquartile range (IQR), resulting in unstable overall performance. In terms of outliers, the lower performance bound of MGSTNet surpasses that of the other deblurring methods. In noise-free experiments, our method achieved the highest Q1 value (30.9293), median value (33.4711), and Q3 value (37.9882). When it comes to deblurring with noise, our approach achieved the highest Q1 (30.5682), median (33.6052), and the second highest Q3 value (37.7561), after MIMO-UNet+ (37.7601). Therefore, MGSTNet outperforms the other methods in deblurring tasks.

\subsection{Experiments on AID, WHU-RS19 Dataset}
To demonstrate the generality of our model, we also applied it to deblur lower-resolution images from the WUH-RS19 dataset. The results for the 19 image categories in WHU-RS19, as well as the average results, can be seen in Table \ref{table6}. Our approach achieves the highest reconstruction performance in terms of the average deblurring value, with an average improvement of 0.34 dB. Moreover, it produces either optimal or sub-optimal deblurring results in most of the image categories. Generally, images with superficial structures (e.g., beaches and deserts) exhibit better deblurring performance, whereas images with more complex structures (e.g., mountains, railway stations, and residential buildings) show slightly lower deblurring performance.

Fig. \ref{fig6} presents a visual comparison of the deblurring results obtained from different methods on the WHU-RS19 dataset. Although other approaches have some success in restoring the blurred input image, they are unable to fully recover local textures and structures in fine detail. GKMNet, which lacks any prior modules, performs worse than the pure network DeblurGAN-v2. Conversely, MIMO-UNet+ and RDN, with their more complex network structures for extracting richer features compared to DeblurGAN-v2, achieve superior deblurring performance. In particular, DeblurGAN-v2, MIMO-UNet, and SFNet introduce noticeable artifacts. On the contrary, our method effectively recovers image details and edges, resulting in the sharpest and most accurate remote sensing images.

\begin{figure*}[htbp]
\centering
\includegraphics[width=7in]{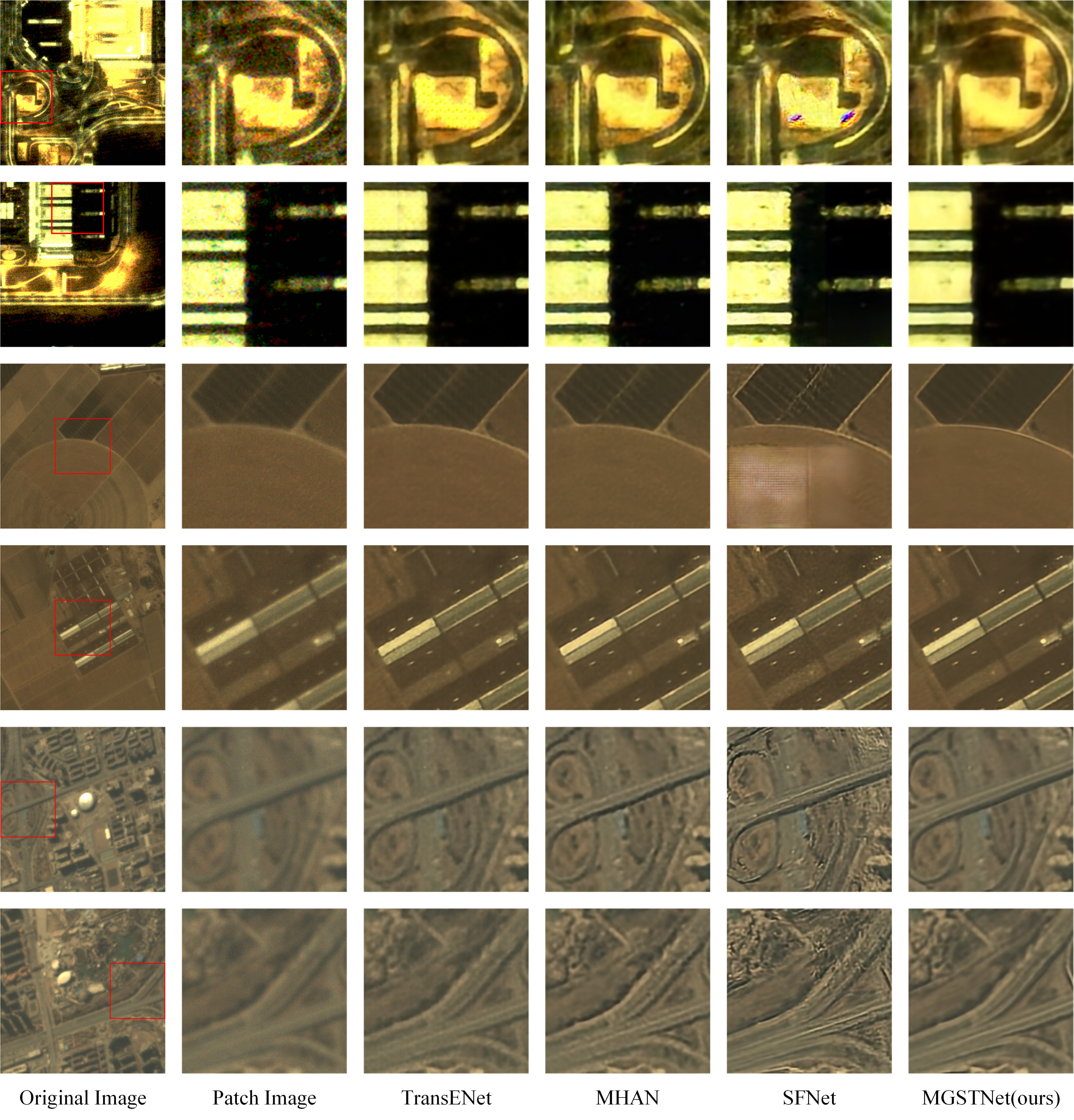}
\caption{Visual comparison of different methods on the Jilin-1 dataset and the Zhuhai-1 datasets.}
\label{fig7}
\end{figure*}

\subsection{Experiments on Real-World Remote Sensing Images}
To evaluate the performance of our proposed method on real remote sensing images, we extracted 12000 images with a size of $256\times256$ from the AIRS data set, where 10000 images are selected as the training set, while the remaining 2000 images are used as the test set. Following the approach used by Wang\cite{Wang_2021_ICCV}, we employed a second-order degradation model to simulate real degradation on synthetic data. Similarly, we employed a sinc filter with a probability of 0.1 to synthesize ringing and overshoot artifacts and generated blur degradation with a probability of 0.9. For blur degradation, we used a combination of Gaussian kernel, generalized Gaussian kernel, and plateau-shaped kernel with probabilities of 0.7, 0.15, and 0.15, respectively. The size of the blur kernel is randomly selected from $\{7, 9, ...21\}$. Additionally, we sampled the standard deviation of the Gaussian blur from the range of 0.2 to 3 (for the second degradation process, the range was 0.2 to 1.5). The shape parameters of the generalized Gaussian and plateau-shaped kernels are sampled from the ranges of 0.5 to 4 and 1 to 2, respectively. To further enhance the realism of the synthetic degraded images, we added Gaussian and Poisson noise with probabilities of 0.5 each. The levels of Gaussian noise and Poisson noise are set to ranges of 1 to 30 and 0.05 to 3, respectively (for the second degradation process, the ranges are 1 to 25 and 0.05 to 2.5). We also introduced a gray noise probability of 0.4. In addition, we applied JPEG compression to the images, using quality factors ranging from 30 to 95. We skipped the second blur degradation with a probability of 0.2. Finally, we applied the sinc filter with a probability of 0.8 after the two degradation processes.

\begin{table}
    \caption{Comparison of different deblurring models on the second-order degraded AIRS dataset.}
    \label{table8}
    \renewcommand\arraystretch{1.1}
    \setlength\tabcolsep{2pt}
    \centering
    \begin{tabular*}{\hsize}{@{}@{\extracolsep{\fill}}lcccc@{}}
    \toprule[1.5pt]
\multicolumn{1}{l}{Method}          & {TransENet} & {MHAN}         & {SFNet}                & {MGSTNet}  \\
\midrule[0.95pt]
{PSNR}                   & {27.26}     & {27.08}        & {25.48}                & {\textbf{27.54}}               \\
{SSIM}                   & {0.7047}    & {0.7003}       & {0.6576}               & {\textbf{0.7176}}              \\
{FSIM}                   & {0.8088}    & {0.7964}       & {\textbf{0.8132}}      & {0.8108}              \\
{VIF}                    & {0.2043}    & {0.2003}       & {0.1816}               & {\textbf{0.2108}}              \\
{IFC}                    & {1.6025}    & {1.5803}       & {1.3907}               & {\textbf{1.6643}}              \\
\bottomrule[1.5pt]
\end{tabular*}
\end{table}

The proposed framework is trained using the training dataset, and its performance is evaluated on both the test dataset and real remote sensing images. The average quantitative comparison of image restoration performance using different filters, such as sinc filtering, isotropic and anisotropic Gaussian, generalized Gaussian, and plateau-shaped blur, on the AIRS test dataset (2000 images) is presented in Table \ref{table8}. Our MGSTNet outperforms other data-driven recovery algorithms in terms of performance. Specifically, compared to recent methods, our algorithm achieves a performance improvement of 0.28 dB over TransENet and a performance improvement of 0.0129 in terms of SSIM.

As shown in Fig. \ref{fig7}, we performed experiments on real remote sensing images collected by the Jilin-1 and Zhuhai-1 datasets. The first and second images are night-light images captured by Jilin-1. We can see that TransENet and SFNet produce noise points and fake color blocks that are not present in the original image, and MHAN generates artifacts along the edges. Compared to other methods, the reconstructed images generated by our method are clearer and have more image details, are closer to the original images, and do not produce false image information. The third and fourth images come from the Jilin-1 dataset, and the fifth and sixth images come from the Zhuhai-1 dataset. We can see that TransENet, MHAN and SFNet all produce obvious distortions and artifacts, whereas our method recovers sharp images.

\subsection{Limitations}
Although our approach has achieved excellent performance in addressing deblurring tasks in remote sensing, there are still limitations to be considered. Firstly, our framework primarily focuses on the degradation models of blur and noise, but the real remote sensing imaging process involves various other factors such as atmospheric turbulence, sensor imaging system degradation, clouds, and haze. These multiple degradations make the restoration process of real images complex and challenging. To tackle this challenge, it is important to incorporate more complex imaging models that account for different combinations of degradations. This will provide guidance for a broader algorithm framework. Secondly, although our method performs well in recovering real images, it may not effectively restore images in extreme scenarios like blooming and overexposure in remote sensing nighttime light images. To address this issue, it is crucial to consider a wider range of situations in the experimental data. This will help to acquire more generalized prior information and improve overall restoration capabilities.

\section{Conclusion}
In this paper, we propose a novel framework for image blind deblurring in remote sensing. Our approach involves bilayer alternating iterations of linear reconstruction and shrinkage thresholds to restore the blurred images. We also provide a theoretical basis for the design of our framework. Additionally, we introduce a multi-scale encoder-decoder architecture that combines feature spaces to learn image feature representations at different scales. This network utilizes nonlinear spatial transformations to extract abstract feature information for multi-scale representations. To enhance the assessment of the blur kernel, we propose a blur kernel proximal mapping module. Furthermore, we employ a multi-scale shrinkage thresholding strategy to ensure a comprehensive representation of the feature space. The results of qualitative and quantitative evaluations on a synthetic remote sensing deblurred dataset demonstrate the effectiveness of our method compared to state-of-the-art approaches. Moreover, we show that our approach is robust in recovering high-quality images even under conditions of blur and noise.

Our future plans involve expanding the capabilities of the MGSTNet to perform more complex and diverse image restoration tasks, such as image super-resolution, image deblocking, image dehazing, and image artifact removal. Additionally, we aim to augment the dataset by incorporating more complex scenario images such as haze images, atmospheric turbulence images, and nighttime light images.
\bibliographystyle{IEEEtran}
\bibliography{DL_Blind_ref_R1}
\vspace{-1cm}
\begin{IEEEbiography}[{\includegraphics[width=1in,height=1.25in,clip,keepaspectratio]{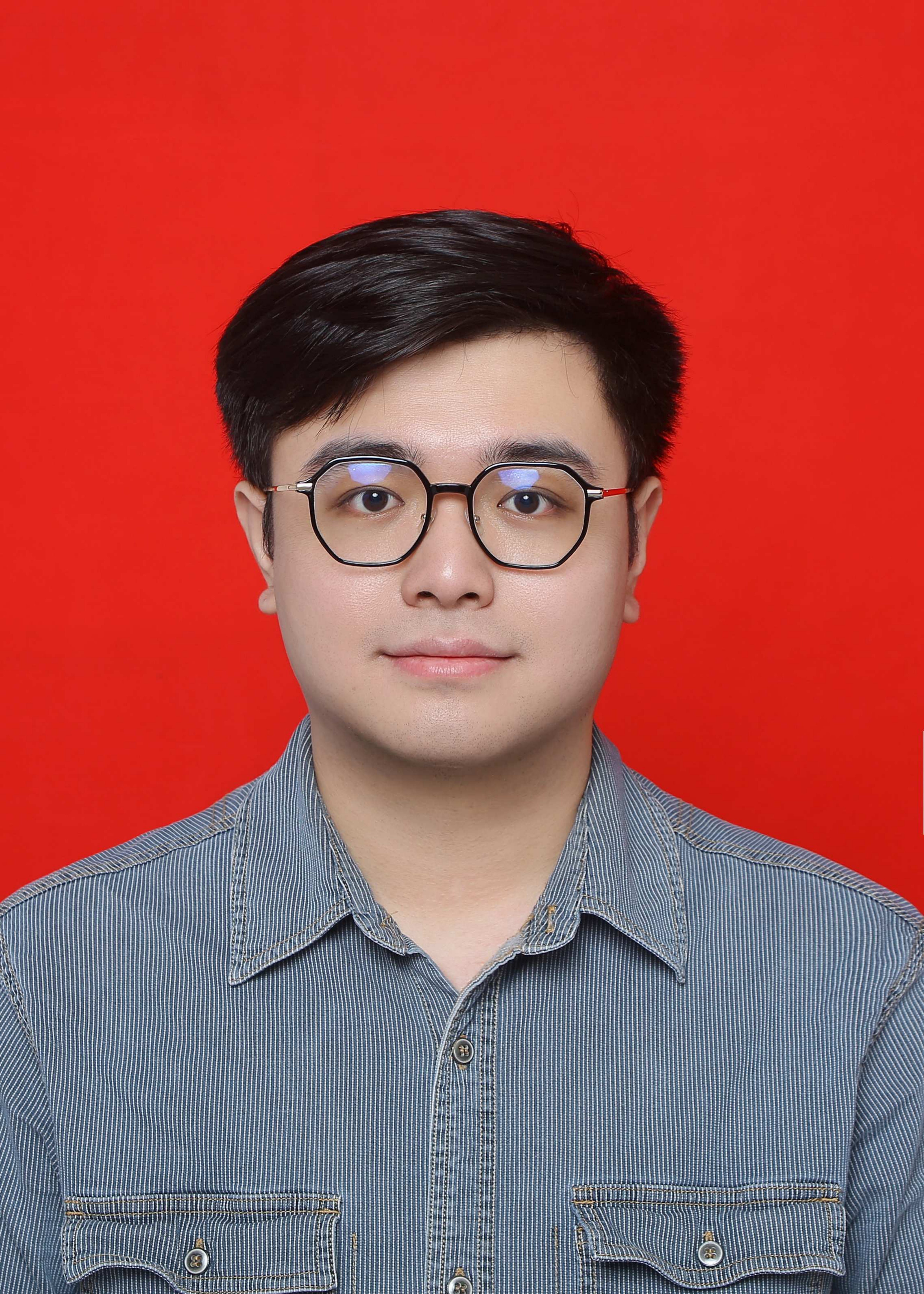}}]{Yujie Feng}
received the B.E. degree from Changsha University, Changsha, China, in 2018. Currently, he is pursuing a Ph.D. degree in Xiangtan University, Xiangtan, China.

His research interests encompass remote sensing image processing.
\end{IEEEbiography}

\vspace{-1cm}
\begin{IEEEbiography}[{\includegraphics[width=1in,height=1.25in,clip,keepaspectratio]{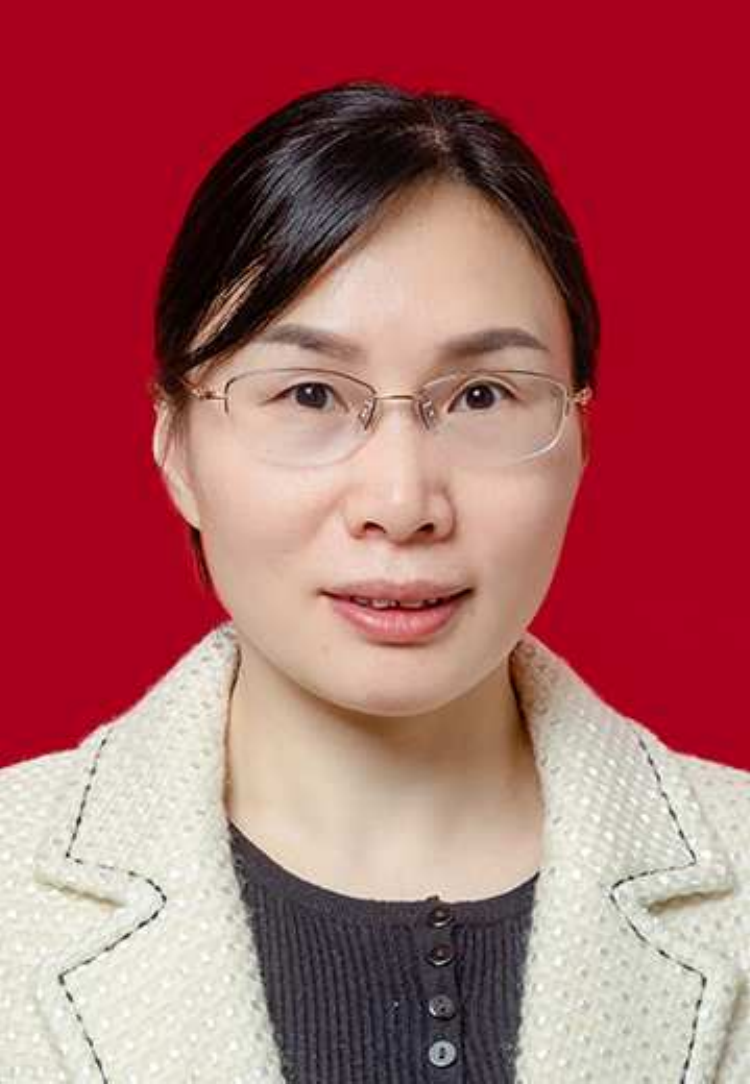}}]{Yin Yang}
received the B.S. degree in Mathematics and Applied Mathematics from Xiangtan University, Xiangtan, China, in 2003, and the M.S. and Ph.D. degrees in Computational Mathematics from Xiangtan University, Xiangtan, China, in 2006 and 2010, respectively.

She is currently a Professor with the Xiangtan University, and the Deputy Director of the National Center for Applied Mathematics in Hunan, Xiangtan. She has presided over the National Natural Science Foundation of China (NSFC), the National Key Research and Development Program, the China-Russia International Cooperation Program of the National Natural Science, the National Natural Science Foundation of China (NSFC), the Hunan Provincial Key Research and Development Program, the Hunan Provincial Outstanding Youth Fund, and the Hunan Provincial Science and Technology Innovation Leadership Program.

Her research interests include applied statistics, image processing, remote sensing data, deep Learning, multiscale model and numerical solution of differential equations.
\end{IEEEbiography}
\vspace{-1cm}
\begin{IEEEbiography}[{\includegraphics[width=1in,height=1.25in,clip,keepaspectratio]{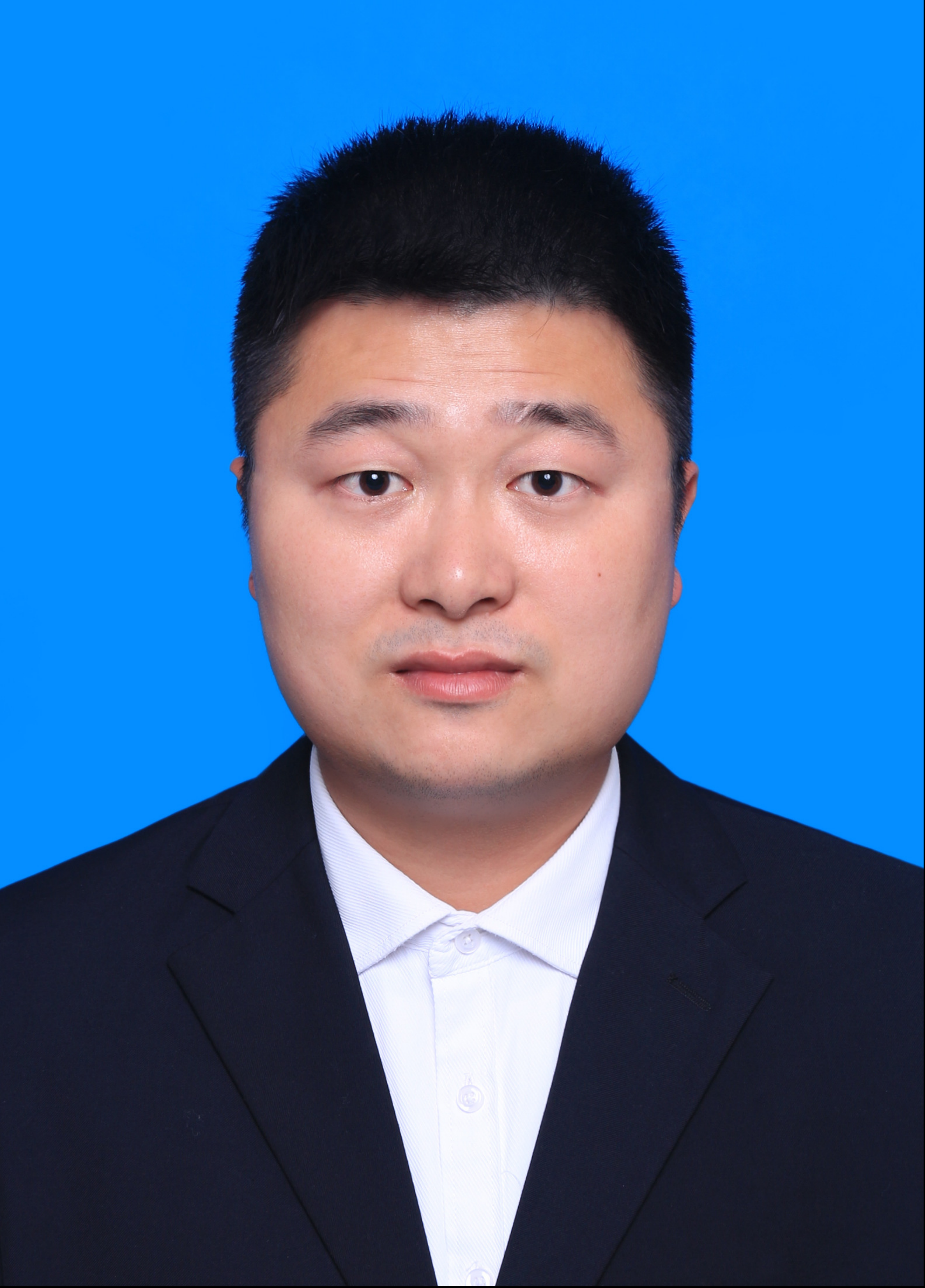}}]{Xiaohong Fan}
received the B.E. degree from Central South University, Changsha, China, in 2014. Currently, he is pursuing a Ph.D. degree in Xiangtan University, Xiangtan, China.

His research interests encompass image reconstruction, image processing, deep learning, optimization and related fields.
\end{IEEEbiography}
\vspace{-1cm}
\begin{IEEEbiography}[{\includegraphics[width=1in,height=1.25in,clip,keepaspectratio]{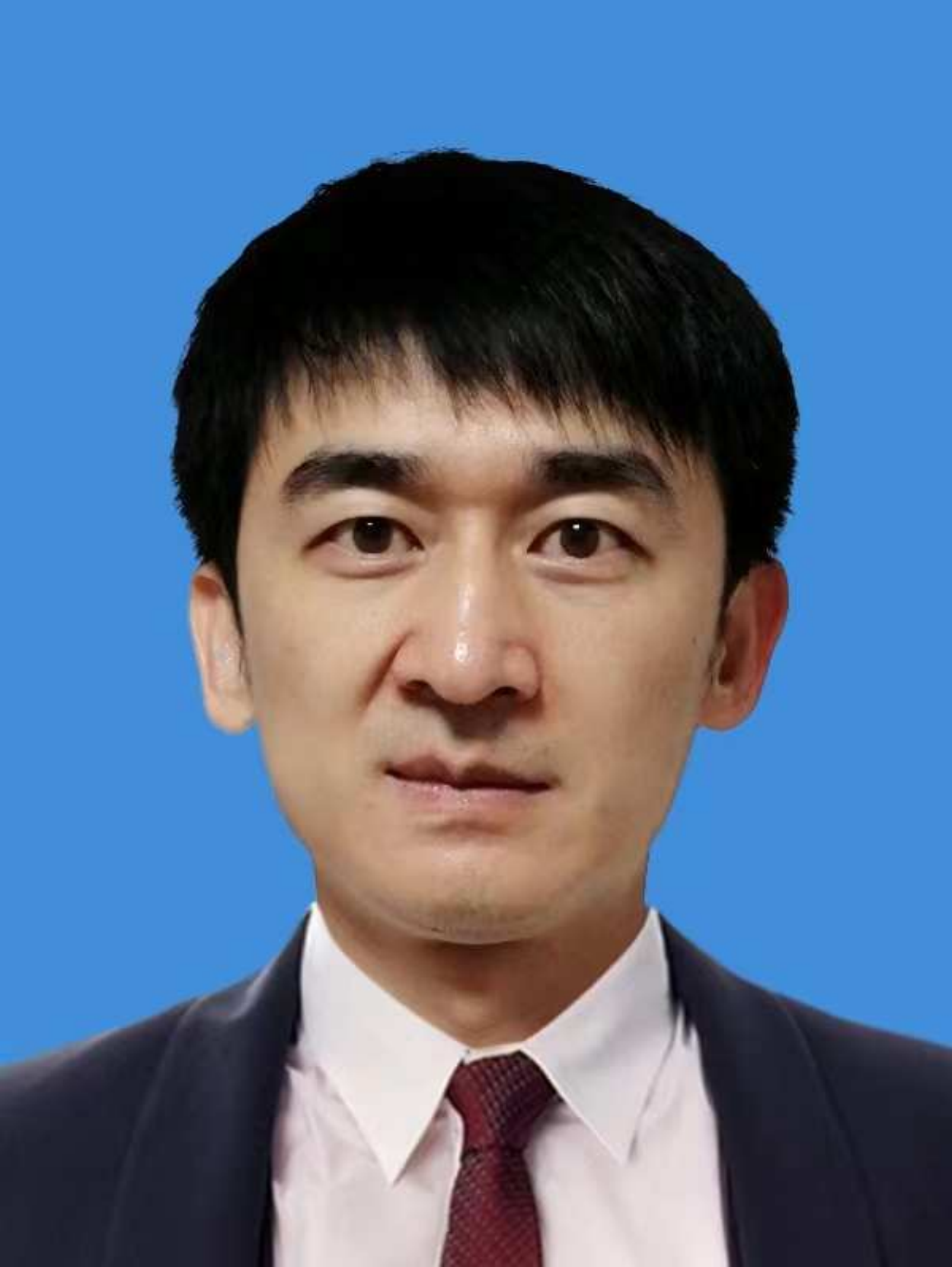}}]{Zhengpeng Zhang}
received the Ph.D. degree from Wuhan University, Wuhan, China, in 2015. He is currently a Full Professor with the School of Automation and Electronic Information, Xiangtan University.

His research interests include remote sensing image processing, navigation and positioning.
\end{IEEEbiography}
\vspace{-1cm}
\begin{IEEEbiography}[{\includegraphics[width=1in,height=1.25in,clip,keepaspectratio]{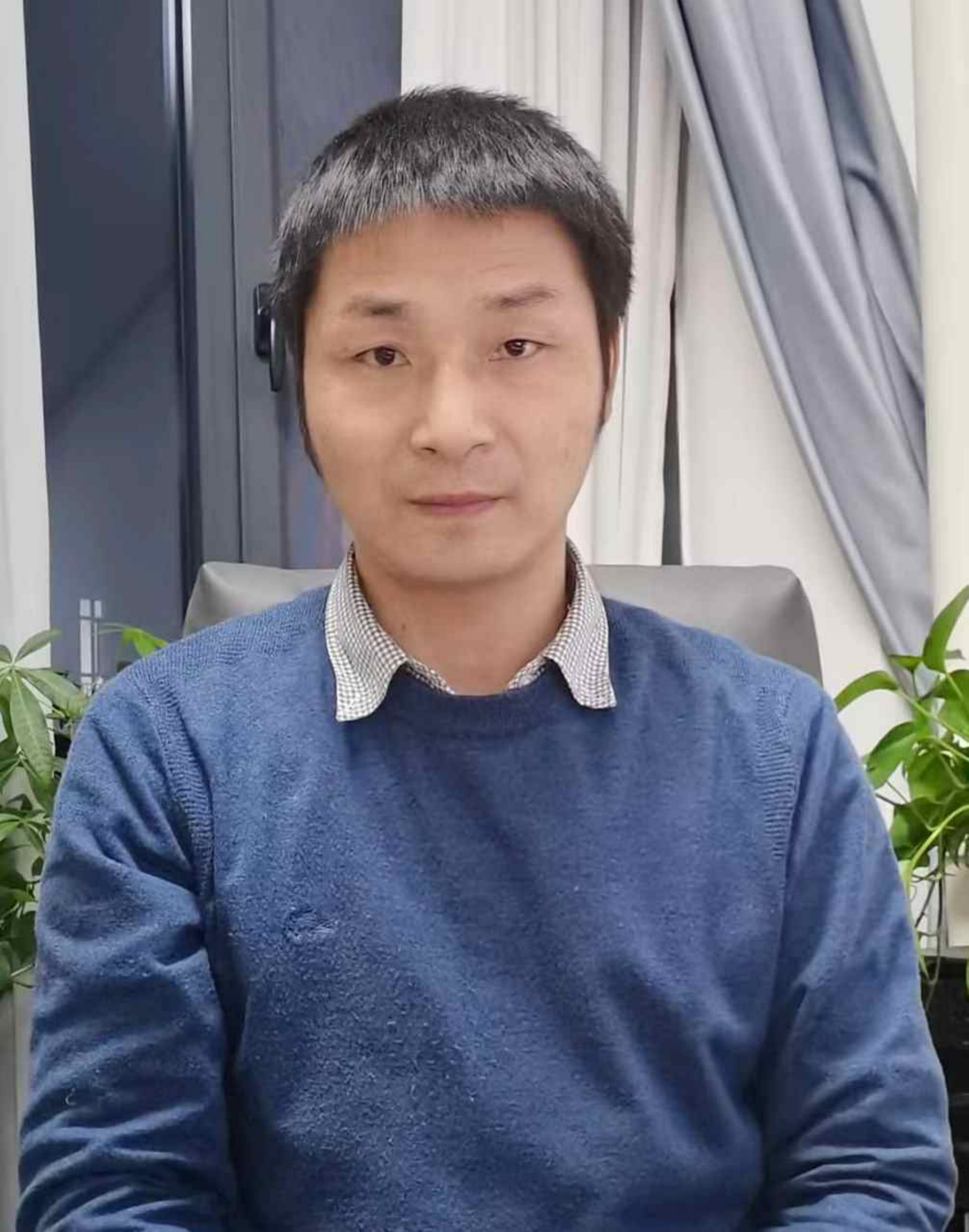}}]{Jianping Zhang}
received the PhD degree from the Dalian University of Technology in 2012. He joined the University of Liverpool as a research associate in 2013. He is currently a full professor with the School of Mathematics and Computational Science, Xiangtan University. 

His current interests are in developing imaging analysis techniques for medical image processing and computer vision problems using a range of mathematical tools such as variational models, PDEs, numerical optimization, and deep learning methods. He has published multiple research papers in leading academic journals, such as SIAM Journals and IEEE Journals. 
\end{IEEEbiography}
\end{document}